\documentclass[10pt,twocolumn,letterpaper]{article}

\usepackage[pagenumbers]{cvpr} 

\definecolor{cvprblue}{rgb}{0.21,0.49,0.74}
\usepackage[pagebackref,breaklinks,colorlinks,allcolors=cvprblue]{hyperref}

\usepackage{appendix}
\usepackage{titletoc}
\usepackage[utf8]{inputenc} 
\usepackage[T1]{fontenc}    
\usepackage{hyperref}       
\usepackage{url}            
\usepackage{booktabs}       
\usepackage{amsfonts}       
\usepackage{nicefrac}       
\usepackage{microtype}      
\usepackage{xcolor}
\usepackage{amsmath}
\usepackage{graphicx}
\usepackage{colortbl}
\usepackage{scalefnt}
\usepackage{pgf}
\usepackage{multirow, makecell, colortbl, hhline}
\usepackage{bbm}
\usepackage{subcaption}
\usepackage{microtype}
\usepackage{animate}
\definecolor{blue-violet}{rgb}{0.44, 0.17, 0.99}
\definecolor{blue-green}{rgb}{0.0, 0.87, 0.87}
\definecolor{dark-green}{rgb}{0.0, 0.67, 0.217}

\def\paperID{7535} 
\def\confName{CVPR}
\def\confYear{2025}

\title{Eval3D: Interpretable and Fine-grained Evaluation for 3D Generation}

\author{
Shivam Duggal*$^1$\quad 
Yushi Hu*$^2$\quad
Oscar Michel$^3$\quad
Aniruddha Kembhavi$^{4}$\quad
William T. Freeman$^1$\\
Noah A. Smith$^{2,5}$\quad
Ranjay Krishna$^{2,5}$\quad
Antonio Torralba$^1$\quad
Ali Farhadi$^{2,5}$\quad
Wei-Chiu Ma$^6$ \\\\
$^1$Massachusetts Institute of Technology\quad 
$^2$University of Washington\quad
$^3$New York University\\
$^4$Wayve\quad
$^5$Allen Institute for AI\quad
$^6$Cornell University
\\
}

\begin{document}

\twocolumn[{%
\renewcommand\twocolumn[1][]{#1}%
\maketitle
\vspace{-10mm}
\captionsetup{type=figure}
\begin{center}
    \includegraphics[width=0.95\textwidth,trim={0cm 0.3cm 0cm 0.2cm},clip]{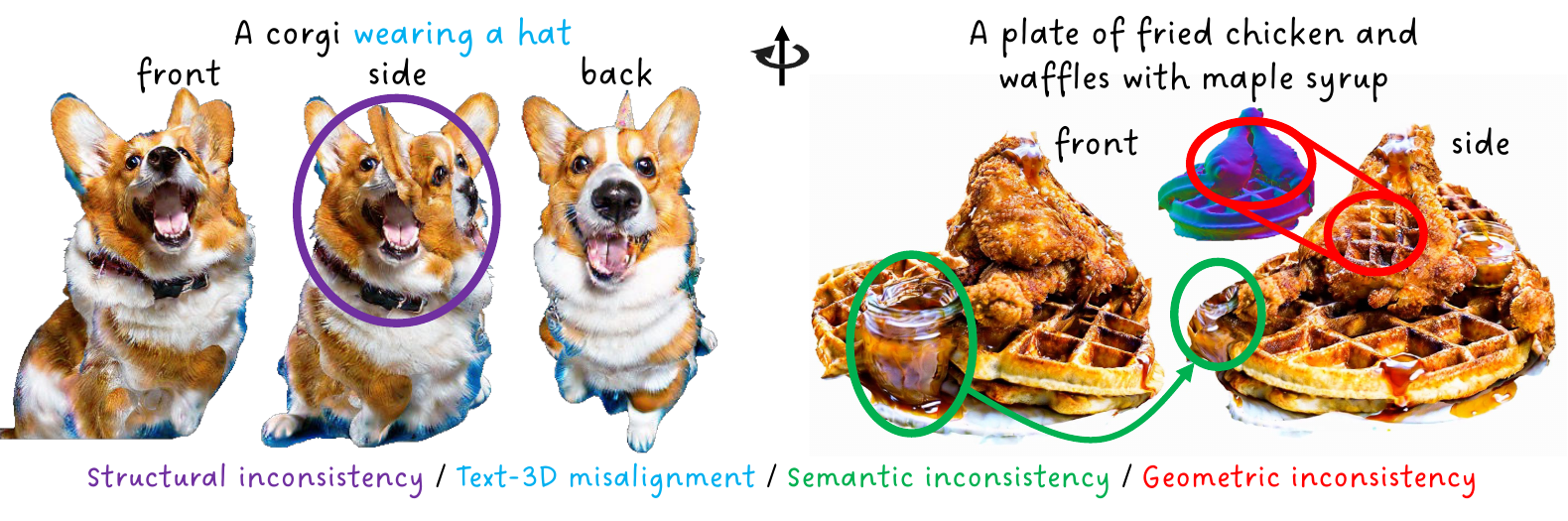}  
\end{center}
\vspace{-7mm}
\captionof{figure}{\textbf{Challenges of 3D generation:} 
    (1) \textcolor{purple}{Structural inconsistency}: lack of globally-coherent 3D shape; (2) \textcolor{cyan}{Text-3D misalignment}: failure to meet the requirements of the input text-prompt; (3) \textcolor{dark-green}{Semantic inconsistency}: content change and incoherent semantics; (4) \textcolor{red}{Geometric inconsistency}: misaligned geometry and texture.}
\label{fig:artifacts}
}
\vspace{3mm}
]

\begin{abstract}
{\vspace{-15mm}\newline}
Despite the unprecedented progress in the field of 3D generation, current systems still often fail to produce high-quality 3D assets that are visually appealing and geometrically and semantically consistent across multiple viewpoints. To effectively assess the quality of the generated 3D data, there is a need for a reliable 3D evaluation tool. Unfortunately, existing 3D evaluation metrics often overlook the geometric quality of generated assets or merely rely on black-box multimodal large language models for coarse assessment. 
In this paper, we introduce Eval3D, a fine-grained, interpretable evaluation tool that can faithfully evaluate the quality of generated 3D assets based on various distinct yet complementary criteria. Our key observation is that many desired properties of 3D generation, such as semantic and geometric consistency, can be effectively captured by measuring the consistency among various foundation models and tools. We thus leverage a diverse set of models and tools as probes to evaluate the inconsistency of generated 3D assets across different aspects. Compared to prior work, Eval3D provides pixel-wise measurement, enables accurate 3D spatial feedback, and aligns more closely with human judgments. We comprehensively evaluate existing 3D generation models using Eval3D and highlight the limitations and challenges of current models. 
Project page: \href{http://eval3d.github.io}{http://eval3d.github.io}.
\end{abstract}

\begin{figure*}[t]
    \centering    
    \vspace{-8mm}    \includegraphics[width=0.85\textwidth,trim={0.7cm 0.6cm 0.cm 0.cm},clip]{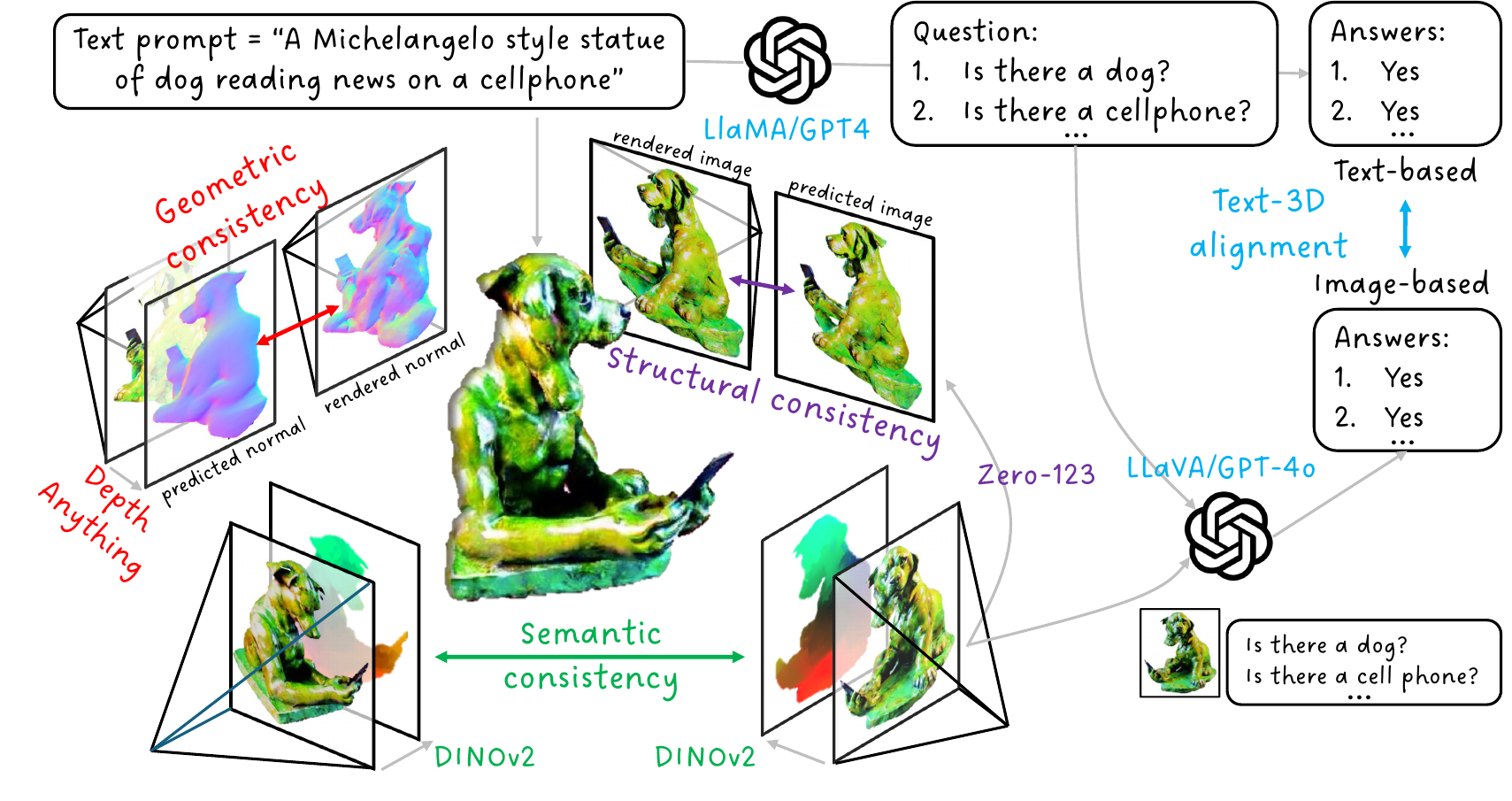} %
    \caption{\textbf{Eval3D} offers interpretable, fine-grained, and human-aligned metrics to assess the quality of 3D generations from various aspects. We utilize a diverse array of foundation models and tools to measure the consistency among different representations of generated 3D assets. 
    }
    \label{fig:method}
    \vspace{-5mm}
\end{figure*}

\vspace{-5mm}
\section{Introduction}
\vspace{-2mm}


High-quality 3D assets play a pivotal role across various domains including gaming, film, augmented/virtual reality, and robotics~\cite{Villapaz2013,yang2024holodeck,JourneytoNanite2022}. 
They empower content creation applications to deliver visually compelling and immersive experiences and enable simulators to realistically reproduce the real world.
Unfortunately, 3D asset creation is complex and tedious, requiring a lot of careful manual effort. Automatic (text/image-to-)3D generation, despite fast progress over the last few years, still falls short in key aspects including geometry, semantic consistency, etc.

\vspace{-3mm}
\paragraph{What makes a good (text/image-to-)3D generation?} 3D assets are not created to merely be rendered onto a 2D image; rather, they are meant to live within a rich 3D world -- the assets may be viewed from various distinct viewpoints, and their 3D geometry will affect how they interact with the surroundings (\emph{e.g.}, collision, reflection, etc). 
Therefore, a high-quality 3D asset should not only be \emph{visually appealing} and \emph{faithful} to its text/image input \emph{across multiple individual viewpoints}, but its \emph{geometry} and \emph{appearance} should also be coherent and plausible \emph{when observed as a whole}.
In the absence of these characteristics, the utility of the generated assets degrades drastically. 
Consider the generated 3D assets in Fig. \ref{fig:artifacts}. 
While the corgi (Fig. \ref{fig:artifacts} (left)) looks reasonable from the front and the back, its overall 3D shape is quite poor and implausible -- notice the Janus issue (\emph{i.e.}, presenting multiple faces). 
Similarly, while the fried chickens and waffles (Fig. \ref{fig:artifacts}(right)) appear faithful from one viewpoint, the 3D asset suffers from semantic and geometric inconsistency across views -- as the object rotates, the maple syrup transform into a jar, and the fried chicken pieces merge with the waffles.  
These 3D-specific structural and consistency challenges 
continue to pose significant difficulties to existing (text/image-to-)3D generation models.

\vspace{-4mm}
\paragraph{What makes a good 3D evaluation metric?} 
To advance (text/image-to-)3D asset generation and to effectively quantify the aforementioned issues, there is a
need for a reliable 3D evaluation tool. 
Unfortunately, while significant progress has been made in evaluating 2D text-to-image models \cite{hu2023tifa,ImageReward,yarom2024you, JaeminCho2024, wiles2024revisiting}, the development of their 3D counterparts is lagging behind. 
Existing 3D evaluation metrics typically rely on black-box multimodal large language models (LLMs), such as GPT-4V, for assessment \cite{Wu2024GPT4VisionIA}, or often overlook the geometric quality of generated assets \cite{He2023T3BenchBC}.
A comprehensive 3D metric should be able to: (a) rate the consistency between the input text/image and the 3D generation and detect misalignments such as a missing object or part, wrong attribute, etc., (b) measure the structural and semantic consistency of the generated asset across the input and the output as well as across multiple views of the 3D asset, and (c) localize flaws and inconsistencies in the produced asset. Such a comprehensive metric would not only permit fine-grained comparisons between different 3D generations but also provide feedback on \emph{where} and \emph{what} to improve. 


\begin{figure*}[t]
    \centering    
    \vspace{-4mm}
    \includegraphics[width=0.85\textwidth,trim={2.7cm 1.5cm 2.7cm 1.5cm},clip]{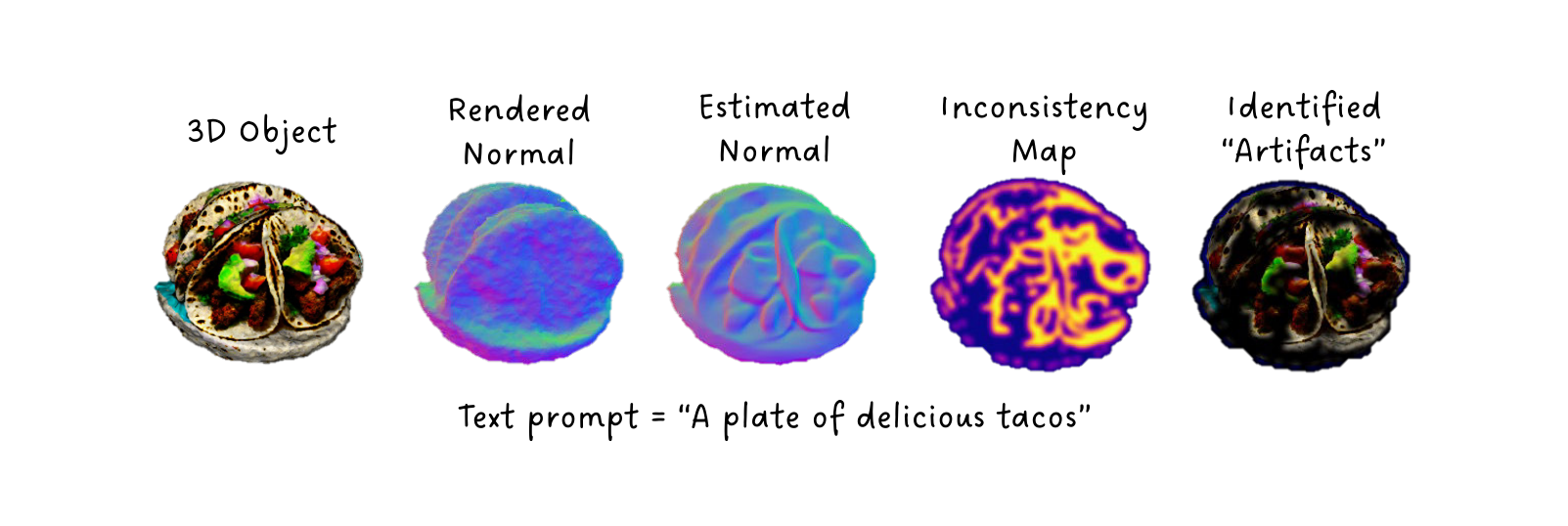} %
    \vspace{-2mm}
    \caption{\textbf{Geometry inconsistency} evaluates texture-geometry misalignment by comparing 3D rendered normal and image-based normal. Bright-yellow indicates large discrepancy. }
    \label{fig:geomtry_inconsistency}
\vspace{-5mm}
\end{figure*}

\vspace{-3mm}
\paragraph{Our method:} We introduce Eval3D, an \emph{interpretable}, and \emph{fine-grained} evaluation tool that measures the quality of generated 3D assets from multiple dimensions.
We build Eval3D based on the observation that many desired properties of 3D generation, such as semantic and geometric consistency, can be effectively captured by measuring the consistency among various foundation models and tools. 
For instance, to measure the geometric consistency of the generated chicken and waffles (\emph{i.e.}, the \textcolor{red}{red} region in \textbf{}Fig. \ref{fig:artifacts}(right)), one can derive the surface normal analytically from the 3D object, and then compare it with the normal estimated by a foundation model (\emph{e.g.}, Depth Anything \cite{depthanything}) from rendered RGB images. 
Based on 2D visual priors, waffles usually possess a grid-like 3D geometry. However, the grid-like patterns do not exist in the normal map derived from 3D. 
By computing the differences, we can automatically identify those regions with inconsistent surface normals and localize where the generative model leverages texture maps to ``cheat.''
We adopt a diverse set of open-source foundation models and tools as probes to evaluate the inconsistency of generated 3D assets across different aspects.
To aggregate the measurements across multiple viewpoints, we further present several simple yet effective 2D and 3D-based fusion mechanisms. 
With these fine-grained, pixel-wise measurements, we can not only easily locate the problematic regions in 2D images (Fig. \ref{fig:geomtry_inconsistency}), but also lift the inconsistency map back to 3D and enable accurate spatial artifact localization (Fig. \ref{fig:3d-inconsistency}). To our knowledge, we are the first to provide such spatial feedback.

To promote the use of our new evaluation tool, we curate a diverse set of prompts and images from multiple sources \cite{threestudio2023,He2023T3BenchBC,dreamcraft3d}.  
We evaluate a variety of (text/image-to-)3D generation models on our data and densely annotate human preferences based on different criteria to verify how well our metrics align with human assessment. 
We show that Eval3D is not only able to automatically detect various generation artifacts (\emph{e.g.}, the Janus issue), but also accurately localize them (Fig. \ref{fig:geomtry_inconsistency}).
Surprisingly, we also find that while existing state-of-the-art generations are visually appealing, they often still suffer from geometric or semantic inconsistencies. 
Finally, we demonstrate that although the accuracy of Eval3D is dependent on the accuracy of the foundation models, it consistently aligns more closely with, or is comparable to, human judgments across all dimensions than prior work. 
As foundation models will improve over time, we hypothesize that Eval3D will become increasingly reliable.

\vspace{-2mm}
\section{Related Work}
\vspace{-1mm}

\paragraph{Generative models for 3D asset generation:} 
Early work on 3D asset generation trained probabilistic generative models directly on 3D data, most commonly using assets from ShapeNet~\cite{chang2015shapenet}. While numerous generative modeling frameworks (\emph{e.g.}, GANs~\cite{gao2022tetgan, gao2022get3d, chen2021decor,kleineberg2020adversarial,wu2016learning,chen2019learning}, VAEs~\cite{gao2022tetgan,dinh2022loopdrawal,Park2019DeepSDFLC,brock2016generative,chen2019learning}, flow models~\cite{cai2020learning,yang2019pointflow}) and 3D representations (\emph{e.g.}, point clouds~\cite{cai2020learning,yang2019pointflow,Nichol2022PointEAS}, meshes~\cite{gao2022get3d,gao2022get3d}, voxels~\cite{chen2021decor,brock2016generative,wu2016learning}, and implicit functions~\cite{chen2019learning,mescheder2019occupancy,Park2019DeepSDFLC,kleineberg2020adversarial}) have been explored, the quality and diversity of 3D generations are constrained by the quantity and characteristics of the available data. 
With this in mind, recent dominant approaches turn to leverage supervision from foundation models trained on massive collections of 2D images~\cite{clip,Saharia2022PhotorealisticTD,Rombach2021HighResolutionIS}. 
During generation, they propagate the gradient from multi-view 2D space back to 3D and iteratively update the underlying 3D representations \cite{Radford2021LearningTV,mildenhall2021nerf,chen2019learning,jain2022zero,michel2022text2mesh,kerbl20233d}. 
Since foundation models encode rich priors of our world, 
this significantly improves the quality of the generated objects. 
To further improve multi-view consistency~\cite{watson2022novel,liu2023zero,shi2023zero123++,zhao2024ctrl123,bourigault2024mvdiff}, object geometry~\cite{magic3d,chen2023fantasia3d,tsalicoglou2023textmesh}, or even the optimization process~\cite{wang2024prolificdreamer,metzer2023latent,yi2023gaussiandreamer}, researchers have proposed a variety of enhancements, such as multi-view diffusion \cite{MVDream}.
While a general trend of progress is evident, the lack of evaluation tools makes it unclear which methods are superior and in which dimensions. Our framework addresses this 
confusion 
by providing a general metric to measure progress, which can be decomposed to identify performance across critical factors relevant to 3D generation.

\vspace{-3mm}
\paragraph{Evaluating 3D generation: } Perhaps closest to our work are GPT-4V~\cite{Wu2024GPT4VisionIA} and T3-Bench~\cite{He2023T3BenchBC}. GPT-4V~\cite{Wu2024GPT4VisionIA} is an Elo-based method that uses the eponymous multimodal LLM to compare the quality of two generated objects, while T3-Bench~\cite{He2023T3BenchBC} employs a vision-language model (VLM) to evaluate the semantic adherence of generated objects to the prompt. 
While we share similar insights, there are several key differences: first,  
unlike GPT-4V, which provides a pairwise preference, Eval3D offers an absolute metric, making it easier to quantify and compare performance across different models and tasks. 
Additionally, both GPT-4V and T3 rely mainly on language outputs for evaluation. The reliance on the inherently imprecise nature of language makes localizing the source of error difficult. 
This issue is particularly problematic for detecting multi-view inconsistencies, as each view may independently seem satisfactory to the VLM. In contrast, our method incorporates semantic judgments from VLMs \cite{hu2023tifa,sun2023dreamsync,cho2024davidsonian} and uses pretrained computer vision models \cite{DINOv2,depthanything} to provide precise and localized estimates of geometric and structural deviations at multiple scales of granularity. These error measurements can be back-projected to 3D, making Eval3D more informative.

\vspace{-2mm}
\section{Eval3D}
\vspace{-2mm}
\label{sec:method}

Our aim is to develop an evaluation tool that can faithfully assess the quality of (generated) 3D assets based on various distinct yet complementary criteria. 
Unlike 2D images, which are merely projections of our visual world, these generated assets live in 3D. 
Their geometry influences how they interact with the environment, and they can be viewed from multiple angles. Therefore, it is crucial to ensure that the appearance of these 3D generations is coherent and visually appealing from various viewpoints, that their geometry is high-quality and consistent with their RGB counterparts, and that the generated content remains faithful to the input conditions (\emph{e.g.}, text, image).

Towards this goal, we devise five distinct measurements to evaluate the quality of generated 3D assets from different aspects. 
Our key observation is that \emph{many of the desired 3D properties can be predicted from one another if the (generated) 3D asset is of high-quality}; 
if the cues are unpredictable, then the 3D generation often has artifacts or is not coherent.
Building upon this intuition, we consolidate a diverse set of foundation models \cite{depthanything,GPT-4o,DINOv2,Stable-Zero123,ImageReward} and tools to compute and predict a wide range of visual cues and measure their consistency. 
The measurements across multiple viewpoints are then aggregated to reflect the quality of the 3D generation with respect to different aspects. 
For each critic, we scale the score to lie within 0-100\%. 

\vspace{-5mm}
\paragraph{Notation:} Let $t$ be the input text prompt and $\mathcal{I^{\text{GT}}}$ be the optional image condition. 
Let $\mathcal{O}_{t}$ be the text-conditioned 3D asset, where the underlying representation could be either a mesh $\mathcal{O}_{t}^{\text{mesh}}$ or an implicit representation $\mathcal{O}_{t}^{\text{imp}}$. Let $\{\mathcal{I}_i\}_{i=1}^N$ denote the rendered RGB images of the generated asset $\mathcal{O}_{t}$ at viewpoint $v_i$. Let $f$ denote the foundation models or tools that we adopt to extract visual cues.

\subsection{Geometric Consistency}

To evaluate how good the geometry of an asset is, one naive way is to consume the 3D asset and output a score. Unfortunately, it is unclear what constitutes a good geometry 
and there is no such data or model that allows us to directly predict the geometric quality. 

In this work, we take a proxy and measure the consistency between the surface normal analytically derived from the 3D asset and the normal predicted by a foundation model from 2D images.  
The intuition is that a good 3D geometry should be reflective of its visual appearance. If the derived geometry does not align with the normal predicted from RGB (\emph{e.g.}, the chicken waffle in Fig. \ref{fig:artifacts}, right), then one of the components is incorrect. 
Specifically, we first compute the analytical surface normal of the 3D asset by taking the normal of each face in the mesh, or by taking the negative gradient of the density \cite{Mildenhall2020NeRF,Boss2020NeRDNR} or signed distance field \cite{NeuS,Yariv2020UniversalDR}, depending on the 3D representation.
Then for each viewpoint, we exploit Depth Anything \cite{depthanything} to compute the monocular depth and convert it into normal. Empirically we find it more robust than direct  surface normal estimation \cite{omnidata}. 
Finally, we project the analytical normal to each view, measure the pixel-wise angular difference, and compute the amount of inliers:
\begin{align}
\text{Geometric} & \text{ consistency} = \nonumber\\ & \frac{1}{N_p}\sum_{p} \mathbbm{1}[\arccos(\mathbf{n}_p^{\text{anal}} \cdot \mathbf{n}_p^{\text{pred}}) < \delta^{\text{norm}}]
\end{align} 
Here, we iterate over all pixels $p$. $\mathbf{n}^{\text{anal}}$ indicates that the normal of pixel $p$ is analytically derived, while $\mathbf{n}^{\text{pred}}$ suggests it is predicted. {${N}_p$ denotes the number of valid rendered pixels.}
In practice, we set the threshold $\delta^{\text{norm}}$ to $23^\circ$ based on a hold-out validation set.  

\subsection{Semantic Consistency}
Another crucial factor for a 3D generation to be appealing and useful is that the underlying content and semantics remain consistent and coherent across different viewpoints. 
Unlike visual appearance, which may be affected by extrinsic factors such as lighting and change across views, the semantics of an asset should be invariant to different perspectives. 
To quantify the degree to which the semantics change across views, we capitalize on large vision foundation models that encode rich semantics. 


We first extract DINO features for every rendered image $\mathcal{F}^{\text{DINO}}_i = f^{\text{DINO}}(\mathcal{I}_i)$. Then, for each 3D point $x$ on the surface of the asset $\mathcal{O}$, we project it onto 2D feature maps $\{\mathcal{F}^{\text{DINO}}_i\}$ to retrieve its corresponding features $\{\mathcal{F}^{\text{DINO}}_i(\pi_{v_i}(x))\}$. Here, $\pi$ is the projection operator. Since a point may be visible from multiple images, we compute the variance of its DINO features. 
Specifically, we calculate the variance along each feature dimension separately across viewpoints and then take the mean of these variances across all feature dimensions. 
Finally, we compute the ratio of vertices whose variance is below a certain threshold.
The intuition is that if the content and semantics of a 3D asset remain relatively stable across different viewpoints, its DINO features should be consistent. 
In practice, for computational efficiency, we only consider the vertices of the 3D mesh $x^{\text{vert}}$ as 3D points:
\begin{align}
\label{eq:semantic_consistency}
\text{Sem}&\text{antic} \text{ consistency} =\\ & \frac{1}{N_{\text{vert}}}\sum_{x^{\text{vert}}} \mathbbm{1}[\text{mean}(\text{Var}(\{\mathcal{F}^{\text{DINO}}_i(\pi_{v_i}(x^{\text{vert}})\})) < \delta^{\text{DINO}}]\nonumber
\end{align}

We set  $\delta^{\text{DINO}}$ to the $70$th percentile of average DINO variance based on a hold-out validation set.

\begin{figure*}[t]
\centering   
\vspace{-4mm}
\includegraphics[width=0.85\linewidth,trim={0.4cm 0.3cm 0.6cm 0.2cm},clip]{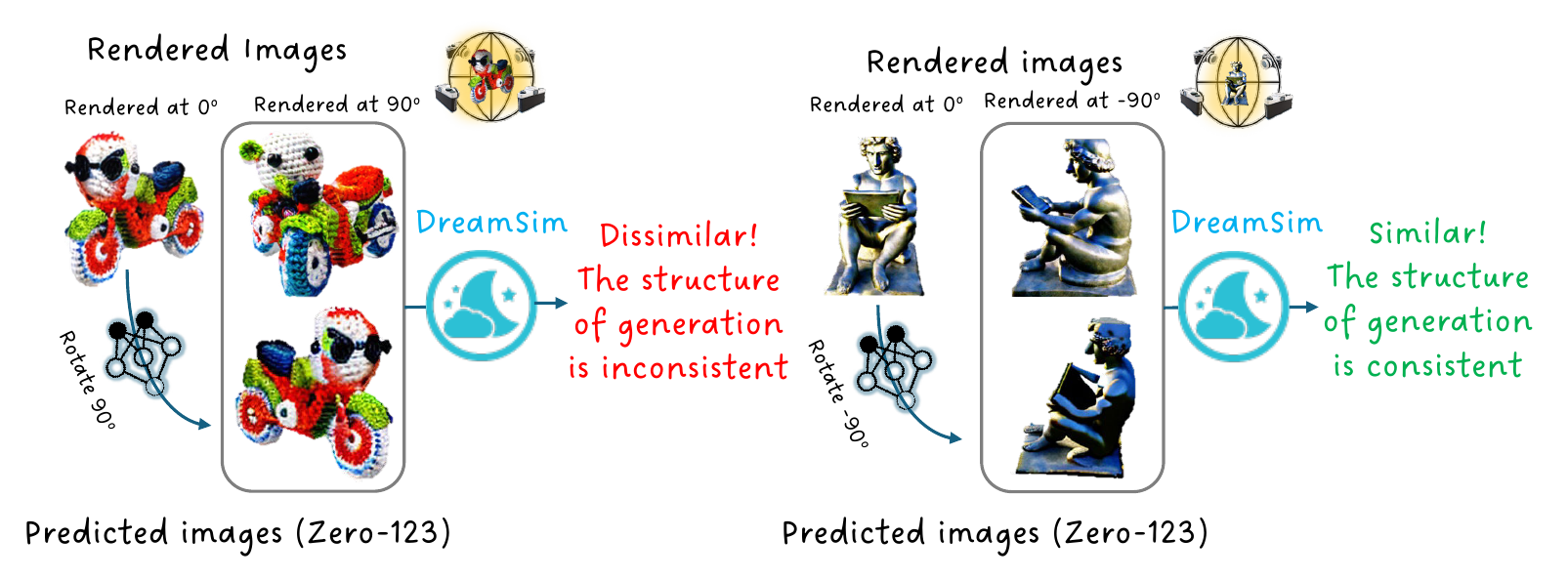} 
\vspace{-2mm}
    \caption{\textbf{Structural consistency} evaluates the  geometric coherence of the generated 3D assets by comparing  rendered views with the predictions from a novel view synthesis model (Zero-123) across various rotations. We utilize DreamSim to assess image similarity.}
    \label{fig:zero123}
\vspace{-5mm}
\end{figure*}

\vspace{-1mm}
\subsection{Structural Consistency}
A generated asset that looks plausible from a single view does not necessarily imply that it remains coherent when observed as a whole. 
The third criterion we are interested in measuring is whether the \emph{overall} structure of the generated asset is coherent and plausible. 

We frame this assessment as a novel view synthesis (NVS) problem. Our intuition is that if the generated asset is overall plausible,  one should be able to synthesize, to some degree, what other views might look like. For instance, given an image of the back of the head, one should be capable of predicting the face, although there might be variations in the size of the eyes or nose compared to the actual generation. 
On the other hand, if the generation suffers from artifacts or noise, it would be difficult to effectively conduct cross-view prediction, and the discrepancies could be significant.

With this motivation in mind, given a rendered RGB image $\mathcal{I}_i$ as input, we first utilize Stable-Zero123 \cite{Stable-Zero123}, a diffusion-based NVS model, to predict what an image from another unobserved viewpoint might look like, denoted as $\mathcal{I}_{i\rightarrow j} = f^{0123}(\mathcal{I}_i; v_i, v_j)$.
We then measure the difference between the predicted image $\mathcal{I}_{i\rightarrow j}$ and the image rendered from the generated asset $\mathcal{I}_{j}$. Since we care more about the perceptual similarity rather than pixel-wise differences, we employ DreamSim \cite{dreamsim}, a human-aligned perceptual metric,
for this similarity assessment (see Fig.~\ref{fig:zero123} for overview). Finally, we repeat the same operation for multiple views $j$ and then take the average. 
In practice, to increase robustness, we consider different images $\mathcal{I}_i$ as input and select the best one:
\begin{align}
\text{Structural} & \text{ consistency} =\nonumber\\ 
& \max_{i} \frac{1}{N}\sum_{j=1}^N \Bigl(1 - f^{\text{DreamSim}}(\mathcal{I}^{\text{pred}}_{i\rightarrow j}, \mathcal{I}_j)\Bigr)
\end{align} 
We take as input both the canonical view (rendered at $0^\circ$), and the rendered image from $90^\circ$. We predict novel viewpoints at an interval of $90^\circ$. 

\subsection{Text-3D Alignment}
Another often overlooked aspect of 3D evaluation is the fidelity of the generated 3D asset to the input conditions. For instance, are all the mentioned objects being generated? Do the relationships among the objects align with what the text describes? The alignment issue is critical as it reflects how accurately the 3D generation adheres to the instructions.

To effectively measure the text-3D alignment, we draw inspiration from TIFA \cite{hu2023tifa}, an evaluation metric for text-to-image generation, and adapt it for assessing text-to-3D generation. 
The key idea is to measure the consistency among large language models (LLMs) and multimodal LLMs across multiple views, and use this consistency as a proxy for evaluation.  
Specifically, we first employ a LLM to generate $M$ multiple-choice question-answer tuples $\{Q_j, C_j, A_j\}_{j=1}^M$ based on the input text prompt $t$, where $Q_j$ is a question, $C_j$ is a set of answer choices, and $A_j \in C_j$ is the gold answer. 
Then, for each question $Q_j$, we employ a multimodal LLM to produce an answer based on each input image 
{$A_{j}^{v_i, \text{VQA}} = f^{\text{MLLM}}(\mathcal{I}_i, Q_j)$. }
If the answer from the multimodal LLM $A_{j}^{v_i, \text{VQA}}$ aligns with the answer derived from the LLM $A_j$, we conclude that the specific property in question is aligned \emph{at such viewpoint} $v_i$. 
For example, given a text prompt $t = \text{``}\texttt{A statue of a dog.}$'', the LLM could generate question-answer tuples such as $\{\text{``}\texttt{Is there a statue?}\text{''}, \{\text{``}\texttt{Yes}\text{''}, \text{``}\texttt{No}\text{''}\}, \text{``}\texttt{Yes}\text{''}\}$, and \emph{view-dependent} alignment is assessed based on whether both the text and the image confirm the presence of a statue (see Fig. \ref{fig:method}). 
To aggregate individual scores, we repeat the same procedure for all questions $\{Q_j\}_{j=1}^M$ and across all viewpoints $\{v_i\}_{i=1}^N$: 
\begin{align}
\text{Text-3D} & \text{ alignment} =\nonumber\\ & \frac{1}{M}\sum_{j=1}^{M} \underset{v_i}{\text{Any}}\bigg( 
 \underset{v \in \text{Adj}(v_i)}{\text{All}} \big(\mathbbm{1}[A_j == A_j^{v, \text{VQA}}]\big) \bigg) 
\end{align} 
A question is considered globally aligned if and only if, for some viewpoint $v_i$, the answers of all its neighboring views are aligned. We find that this method is effective for dealing with occlusions and the noise of multimodal LLM.
In practice, we adopt LLaVA-NeXT-7B~\cite{li2024llavanext-strong} as our VQA model, and we follow DSG~\cite{JaeminCho2024} for generating and evaluating our question-answer pairs.

\vspace{-1mm}
\subsection{Aesthetic}
Finally, given that a generated 3D asset can be viewed from a wide range of angles, it is crucial to ensure that it appears visually appealing across \emph{all} possible views. 
In this paper, we exploit two different approaches, with varying  computational cost, to compute the aesthetic score $f^{\text{aesthetic}}(\{\mathcal{I}_i\}).$
First, following \cite{Wu2024GPT4VisionIA}, we use  GPT-4o~\cite{GPT-4o} to conduct pairwise quality comparisons of 3D assets and compute the ELO score of each model.
We improve the prompts and normalize the ELO score to 0-100\% (by comparing with 3D generation models in Sec.~\ref{sec:exp:details}).
While effective, the closed-source nature and the high computational cost resulting from combinatorial comparison may create barriers to large-scale evaluation.
As an alternative, we employ ImageReward \cite{ImageReward}, an open-source aesthetic estimator, to measure the aesthetic quality of each rendered image $f^{\text{ImgReward}}(\mathcal{I}_i)$ and then compute the average.
We compare both implementations in Sec. \ref{sec:exp} and discuss their respective strengths.


\vspace{-2mm}
\subsection{Localizing 3D Artifacts} 
\begin{figure}[t]
    \centering 
    \vspace{-3mm}
    \includegraphics[width=0.9\linewidth,trim={1.3cm 0.cm 0.4cm 0.1cm},clip]{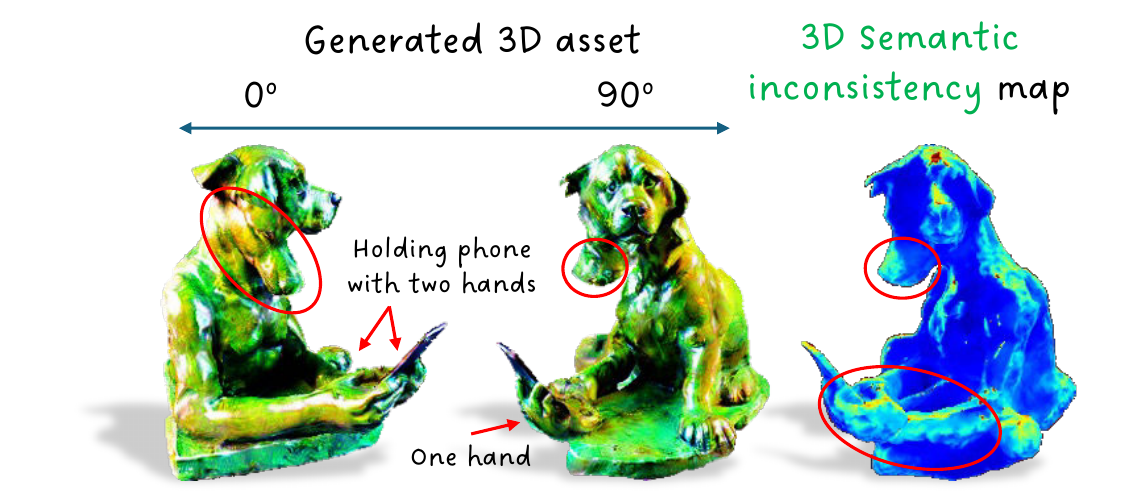} %
    \vspace{-2mm}
    \caption{\textbf{3D inconsistency maps:} The proposed 3D metrics, semantic and geometric consistencies, allow fine-grained localization of the artifacts (eg: Janus issue: mutliple nose / face, inconsistent hand geometry, arbitrary surface patterns on the back) by fusing / computing the metrics in 3D space.}
    \label{fig:3d-inconsistency}
\vspace{-4mm}
\end{figure}

Unlike previous works \cite{He2023T3BenchBC, Wu2024GPT4VisionIA}, our proposed geometric and semantic consistency measures operate in the 3D space, allowing us to not only classify the generation as an artifact but also to \textit{localize the artifacts} on the extracted 3D mesh. For the semantic consistency 
, artifacts can be localized by highlighting outlier vertices—specifically, mesh vertices with DINO feature variance exceeding a certain threshold. Fig.~\ref{fig:3d-inconsistency} illustrates an example where vertices with high semantic inconsistency (Eq.~\ref{eq:semantic_consistency}) significantly contribute to the generated 3D inconsistencies like the Janus issue and extraneous geometries. For the geometric consistency,
we back-project the pixel-wise cosine distance map between the predicted and rendered geometric normals onto the 3D mesh, highlighting vertices where the mean or maximum cosine distance across different viewpoints exceeds a threshold. Finally, as shown is 
Fig.~\ref{fig:3d-inconsistency}, fusing the proposed semantic or geometric consistency metrics to the 3D space could enable estimating 3D uncertainty at novel viewpoints. Such consistency cues can be leveraged as a self-supervised refinement metric to enhance the generated geometry.


\begin{table*}
\centering
\vspace{-3mm}
\scalebox{0.9}{
\begin{tabular}{lccccc}
\specialrule{.2em}{.1em}{.1em}
\rowcolor{gray!5} Approach & \multicolumn{1}{c}{Geometric Consis. $\uparrow$} & \multicolumn{1}{c}{Semantic Consis. $\uparrow$} & \multicolumn{1}{c}{Structural Consis. $\uparrow$} & {Aesthetics $\uparrow$} & \multicolumn{1}{c}{Text-3D Align. $\uparrow$}\\
\specialrule{.1em}{.05em}{.05em} \vspace{-2mm}\\
CLIP-Score \cite{hessel2022clipscore} & 43.6 & - & 64.0 & 62.0 & 74.1 \\
ImageReward \cite{ImageReward} & 44.3 & - & 63.7 & 75.4 & 79.5 \\
T3-Bench \cite{He2023T3BenchBC} & - & - & - & 76.5 &  56.7\\
GPT4V \cite{Wu2024GPT4VisionIA} & 46.9 & - & 68.9 & 85.6 & 72.8 \\
\rowcolor{green!5} Eval3D {(Ours)} & \textbf{83.0} & \textbf{68.0} & \textbf{69.2} & \textbf{87.4} & \textbf{88.7} \\
\hline
\end{tabular}
}
\vspace{-2mm}
\caption{\textbf{Pairwise rating agreements between human and evaluation metrics.} We measure the probability that the automatic evaluation makes the same decision as human annotators when comparing two 3D assets. Eval3D achieves the best alignment with humans.}
\label{tab:text-3D-human-alignment}
\vspace{-3mm}
\end{table*}

\vspace{-2mm}
\section{Experiments}
\vspace{-2mm}
\label{sec:exp}

In this section, we begin by discussing the curated Eval3D evaluation benchmark. (Sec.~\ref{sec:eval3d_benchmark})
We then outline the experimental details, covering the targeted 3D generative models, experimental setup, and human annotation process.(Sec.~\ref{sec:exp:details})
Finally, we present our experimental results. (Sec. ~\ref{sec:exp:results})
We first assess the agreements between Eval3D and human annotations, showing that Eval3D aligns well with humans on all evaluation dimensions.
Then we benchmark text/image-to-3D algorithms with Eval3D and discuss our findings.

\vspace{-0.5mm}
\subsection{Eval3D Benchmark}
\label{sec:eval3d_benchmark}
\vspace{-0.5mm}

To thoroughly evaluate 3D generation algorithms, we curate the Eval3D benchmark. Unlike prior work, which provides only text prompts, Eval3D additionally includes a scene graph, dense human annotations, and optionally a carefully selected SDXL image suitable for image-to-3D generation.


\vspace{-3mm}
\paragraph{Prompt source:} Eval3D contains 160 text prompts for text-to-3D evaluation. The prompts come from diverse sources, including threestudio~\cite{threestudio2023} and the GPT-generated prompts from prior works~\cite{dreamcraft3d, Wu2024GPT4VisionIA, He2023T3BenchBC}. We curate the prompt source such that 80 prompts contain a single object, and 80 prompts contain multiple objects. For image-to-3D evaluation, we use the 20 image-text pairs from Dreamcraft3D~\cite{dreamcraft3d}.

\vspace{-3mm}
\noindent\paragraph{Scene graphs:}
To facilitate further investigation, we also provide the scene graph for each prompt. 
Following prior work~\cite{JaeminCho2024}, the scene graph contains 5 kinds of elements: entity (\emph{e.g.}, ``statue''), attribute (\emph{e.g.}, colors, materials), relation (\emph{e.g.}, ``on top of''), action (\emph{e.g.}, ``sit''), and others (\emph{e.g.}, art style). Please refer to Appendix for more details.

\vspace{-3mm}
\noindent\paragraph{Annotations:}
We  provide dense expert annotations on all 3D evaluation criteria discussed in Sec.~\ref{sec:method} for 160 prompts across six 3D generation models. We note that our annotation is \emph{an order of magnitude} larger than prior work. For example, \cite{Wu2024GPT4VisionIA} only contains pairwise comparison for 3 prompts for each model pair. \cite{He2023T3BenchBC} annotates quality and alignment for 90 prompts, ignoring geometric, semantic, and structural consistency. See Sec.~\ref{sec:exp:details} and Appendix for more details.

\begin{figure}[t]
    \centering 
    \vspace{-5mm}
    \animategraphics[autoplay,loop,width=0.9\linewidth, trim={0 0 0 0}, clip]{10}{figs/qual-animation/}{001}{034}    
    \vspace{-2mm}
    \caption{\textbf{Evaluation examples from Eval3D.} While the generation from MV-Dream appears visually appealing at first glance, both the boat and the paddle are missing (reflected in \textcolor{cyan}{text-3D alignment}). There are also geometric artifacts on the left side of the panda (reflected in \textcolor{red}{geometric consistency}). Please refer to the Appendix for detailed breakdown of each metric. To view the \textbf{embedded videos}, we recommend using \textbf{Adobe Acrobat Reader}. }
    \label{fig:eval3d-qual}
\vspace{-4mm}
\end{figure}

\vspace{-0.5mm}
\subsection{Experimental Details}
\label{sec:exp:details}
\vspace{-0.5mm}

\paragraph{Target 3D generation models:} We evaluate the proposed metrics on six state-of-the-art text-to-3D algorithms (\emph{i.e.}, DreamFusion \cite{dreamfusion}, Magic3D \cite{magic3d}, ProlificDreamer \cite{prolificdreamer}, TextMesh \cite{TextMesh}, and MVDream \cite{MVDream}) and two image-to-3D algorithms (\emph{i.e.}, Magic123 \cite{Magic123} and DreamCraft3D \cite{dreamcraft3d}) whose implementations are available on ThreeStudio \cite{threestudio2023}. 
These algorithms represent a diverse set of text/image-to-3D generative models, featuring different 3D representations (NeRF, NeRF + Tetrahedral Mesh, Gaussian Splatting), various optimization strategies (single-stage vs. multi-stage), and different generative (diffusion) priors (single view vs. multi-view).
Consequently, the generated assets exhibit a wide range of artifacts, enabling a thorough study of fine-grained 3D evaluation. 
Please refer to Appendix for details.

\vspace{-3mm}
\noindent\paragraph{Setup:}

We render each generated 3D asset from 120 different viewpoints, with cameras positioned at an elevation of $15^\circ$ and an azimuth varying from $0^\circ$ to $360^\circ$ degrees. We render RGB images, surface normals, and opacity maps at a resolution of $512 \times 512$. To compute semantic consistency, we extract the underlying 3D mesh using either marching cubes \cite{marching_cubes} or DMTet \cite{shen2021dmtet}. We upsample DINO features to $256 \times 256$ using FeatUp \cite{fu2024featup} and then fuse them with the extracted 3D mesh. Due to cost considerations, we use 12 viewpoints for the text-3D alignment metric.

\vspace{-3mm}
\noindent\paragraph{Human annotations:} 

Previous human annotation data either lacked coverage for all 3D evaluation dimensions (\emph{e.g.}, only focuses on aesthetics and alignment, ignoring geometry and structure) or included a very limited number of annotations.
To address these issues, we performed dense expert annotations on \emph{all} 3D assets generated by six text-to-3D models based on different criteria discussed in Sec. \ref{sec:method}. 
We carefully designed the annotation process for each metric to ensure high inter-annotator agreement. 
Take \textbf{geometric consistency} for example. For each text prompt, we showed annotators videos of RGB and rendered surface normals, placed side-by-side for all six algorithms. Then we asked annotators to rank the generations based on how well the texture aligns with the surface normals. Ties were allowed. 
We adopted a similar annotation procedure for \textbf{aesthetics}, but showing only the RGB videos. 
As for \textbf{semantic and structural consistency}, since it is difficult to rank (\emph{e.g.}, whether a two-faced dog is better or worse than a three-legged astronaut), we instead framed it as a yes/no question and asked annotators whether they observed any inconsistencies.
Finally, for \textbf{text-3D alignment}, annotators looked at the RGB videos and answered a series of multi-choice questions generated automatically from the text prompt, as discussed in~\cite{JaeminCho2024}.
We direct the readers to the Appendix for more comprehensive details on our annotation process.

\vspace{-0.5mm}
\subsection{Experimental Results}
\label{sec:exp:results}
\vspace{-0.5mm}

\paragraph{Human alignment:} 
An optimal 3D evaluation metric should align closely with human judgment. We assess this alignment by comparing human annotations with ours and baseline metrics. As shown in Tab.~\ref{tab:text-3D-human-alignment}, Eval3D outperforms all baseline evaluation metrics across all dimensions. 
Specifically, our text-3D metric aligns approximately \textbf{16\%} better with human annotations than GPT-4V \cite{Wu2024GPT4VisionIA}. 
Our geometric consistency metric, which is analogous to GPT-4V's texture-geometry alignment \cite{Wu2024GPT4VisionIA}, achieves \textbf{36\%} better alignment with human judgments compared to all other baselines. Importantly, our geometric and semantic consistency metrics are also interpretable and fine-grained, allowing us to localize potential artifacts (see Figs.~\ref{fig:3d-inconsistency} and~\ref{fig:geomtry_inconsistency}). 
We note that we are the first to capture the multi-view semantic consistency.

\begin{table*}
\centering
\vspace{-4mm}
\scalebox{0.9}{
\begin{tabular}{lcccccc}
\specialrule{.2em}{.1em}{.1em}
\rowcolor{gray!5} Algorithm & \multicolumn{1}{c}{Geometric Consis. $\uparrow$} & \multicolumn{1}{c}{Semantic Consis. $\uparrow$} & \multicolumn{1}{c}{Structural Consis. $\uparrow$} & {Aesthetics $\uparrow$} & \multicolumn{1}{c}{Text-3D Align. $\uparrow$}\\
\specialrule{.1em}{.05em}{.05em} \vspace{-2mm}\\
TextMesh~\cite{TextMesh} & \cellcolor{green!10}{83.1} & \cellcolor{green!20}{74.1} & {69.6} & {18.9} & {43.2}  \\
DreamFusion~\cite{dreamfusion} &  \cellcolor{green!30}{88.3} & \cellcolor{green!10}{67.1} & \cellcolor{green!10}{77.6} & \cellcolor{green!5}{24.2} & \cellcolor{green!5}{50.8} \\
Magic3D~\cite{magic3d} & \cellcolor{green!40}{94.1} & \cellcolor{green!5}{61.5} & \cellcolor{green!20}{79.3} & \cellcolor{green!10}{51.2} & \cellcolor{green!10}{64.2} \\
ProlificDreamer~\cite{prolificdreamer} & \cellcolor{green!5}{77.1} & {59.5} & \cellcolor{green!5}{75.1} & \cellcolor{green!30}{60.6} & \cellcolor{green!40}{76.9} \\
Gaussian Splatting~\cite{yi2023gaussiandreamer} & {67.9} & \cellcolor{green!40}{84.8} & \cellcolor{green!40}{81.8} & \cellcolor{green!20}{62.3} & \cellcolor{green!20}{66.7} \\
MVDream~\cite{MVDream} & \cellcolor{green!20}{85.8} & \cellcolor{green!30}{74.3} & \cellcolor{green!30}{81.1} & \cellcolor{green!40}{82.5} & \cellcolor{green!30}{67.7} &  \vspace{1mm} \\
\hline
\end{tabular}
}
\caption{\textbf{Eval3D results of Text-to-3D models.} We benchmark the text-to-3D models discussed in Sec.~\ref{sec:exp:details} on all the evaluation criteria in Eval3D. For each criterion, the score ranges from 0 to 100\%. Higher values indicate better performance.}
\label{tab:text-to-3D-comparison}
\end{table*}

\begin{table*}
\centering
\scalebox{0.9}{
\begin{tabular}{lcccccc}
\specialrule{.2em}{.1em}{.1em}
\rowcolor{gray!5} Algorithm & Input & \multicolumn{1}{c}{Geometric Consis. $\uparrow$} & \multicolumn{1}{c}{Semantic Consis. $\uparrow$} & \multicolumn{1}{c}{Structural Consis. $\uparrow$} & {Aesthetics $\uparrow$} & \multicolumn{1}{c}{Text-3D Align. $\uparrow$}\\
\specialrule{.1em}{.05em}{.05em} \vspace{-2mm}\\

Magic3D~\cite{magic3d} & & \cellcolor{green!20}{97.1} & \cellcolor{green!0}{67.2} & \cellcolor{green!0}{80.3} & \cellcolor{green!0}{25.9} & \cellcolor{green!20}{90.2} \\

MVDream~\cite{MVDream} & &  \cellcolor{green!0}{89.7} & \cellcolor{green!20}{76.1} & \cellcolor{green!40}{83.4} & \cellcolor{green!40}{76.8} & \cellcolor{green!0}{80.5} \\
Magic123~\cite{Magic123} & \checkmark & \cellcolor{green!40}{97.6} & \cellcolor{green!10}{70.7} & \cellcolor{green!20}{82.0} & \cellcolor{green!10}{39.2} &  \cellcolor{green!10}{88.4}\\

DreamCraft3D~\cite{dreamcraft3d} & \checkmark & \cellcolor{green!10}{97.0} & \cellcolor{green!40}{82.8} & \cellcolor{green!10}{81.3} & \cellcolor{green!20}{69.1} & \cellcolor{green!40}{91.9}\\
\hline
\end{tabular}
}
\caption{\textbf{Eval3D results of image-to-3D models.} We evaluate Magic123~\cite{Magic123} and DreamCraft3D~\cite{dreamcraft3d}, both of which take images and texts as inputs. We also show the results of two text-to-3D models (Magic3D~\cite{magic3d}) and MVDream~\cite{MVDream}) using the same text prompts as inputs.}
\label{tab:text-image-to-3D-comparison}
\vspace{-4mm}
\end{table*}

\vspace{-3mm}
\noindent\paragraph{Benchmarking text-to-3D models:} 
Tab.~\ref{tab:text-to-3D-comparison} shows the scores (scaled to $0-100$\%) of each text-to-3D model under various criteria. \textbf{MVDream~\cite{MVDream} generally performs the best,} ranking 1st or 2nd on all metrics, except for geometry consistency.
Upon inspecting the MVDream generations, we find that Eval3D successfully captures its weaknesses: despite their high overall quality, MVDream assets suffer from holes and freqently hallucinate details. 
\textbf{DreamFusion \cite{dreamfusion}, TextMesh \cite{TextMesh}, and Magic3D \cite{magic3d} perform poorly across most metrics.} They produce low-resolution geometries, which result in low aesthetic scores. Additionally, they do not accurately adhere to the text prompts, achieving only 40\%-65\% in text-3D alignment.
Nevertheless, their geometric consistency is comparable to, or better than, that of MVDream. We hypothesize that this is due to their tendency to generate simpler and more convex geometries. 
\textbf{ProlificDreamer achieves high scores in aesthetics and text-3D alignment, yet it ranks last or second-to-last in all three geometric metrics.} In contrast, previous studies \cite{Wu2024GPT4VisionIA,He2023T3BenchBC} rate ProlificDreamer highly.
Upon inspecting these outputs, we observe that Eval3D more effectively highlights the weaknesses of ProlificDreamer. Fig~\ref{fig:3d-inconsistency} showcases a generation by ProlificDreamer that, despite its aesthetic sharpness and alignment with the text prompt, suffers from geometric, structural, and semantic inconsistencies. 
\textbf{Gaussian Splatting ranks the lowest on geometric consistency.} This validates~\cite{Huang2DGS2024}'s finding that Gaussian-Splatting-based methods generate extremely noisy geometry. Interestingly, despite the noisy geometries, the generations rank highly in terms of semantic consistency measured by DINOv2 \cite{DINOv2} and multi-view structural consistency measured using Zero123 \cite{Zero123}. 
We show a few qualitative examples as well as their evaluation results in Fig. \ref{fig:eval3d-qual}. 
Please refer to the Appendix for those of the other text-to-3D algorithms.

\vspace{-3mm}
\noindent\paragraph{Benchmarking image-to-3D models:} 
Unlike previous studies \cite{He2023T3BenchBC, Wu2024GPT4VisionIA}, we also evaluate state-of-the-art image+text-to-3D generative algorithms (\emph{i.e.}, Magic123\cite{Magic123} and DreamCraft3D\cite{dreamcraft3d}). 
To validate to which degree an input image could help, we compare these algorithms with two text-only 3D generation algorithms, Magic3D~\cite{magic3d} and MVDream \cite{MVDream}.
As shown in Tab.~\ref{tab:text-image-to-3D-comparison}, text+image-to-3D generations better align with the text prompts and perform comparable on geometric and structural consistency metrics. 
In particular, although Magic123 and Magic3D share similar generation pipelines, Magic123, with the aid of image guidance, achieves improved semantic and structural consistency. We conjecture that the improvement stems from the use of diffusion models to generate novel views from the input image, which serve as priors and help avoid structural inconsistencies like the Janus problem.

\section{Analysis and Discussion}
\label{sec:discussion}
\paragraph{Open-source vs.~closed-source evaluation:} 
Compared to previous work \cite{Wu2024GPT4VisionIA}, which exclusively capitalized on closed-source multimodal LLMs for assessment, we propose a fully open-source solution for 3D evaluation. 
Additionally, we design Eval3D to be flexible, allowing foundation models and tools to be easily swapped with alternatives possessing similar capabilities. 
For instance, one could replace the open-source LLaVA model \cite{li2024llavanext-strong} with the closed-source GPT-4o for the text-3D alignment metric, or alternatively, select either GPT-4o or ImageReward \cite{ImageReward} for aesthetic measurement. 
We experimented with these models and found that for text-3D alignment, our human alignment is similar using two models (88.7\% using LLaVA, 91.5\% using GPT-4o), all substantially outperforming  prior metrics. While GPT-4o aligns more closely with human judgments than ImageReward does, the overall ranking of 3D algorithms assessed by the two is similar, with an edit distance of 1. These results suggest that open-source models are effective for evaluation.

\vspace{-3mm}
\noindent\paragraph{Limitations:}
The foundation models used in the paper are not perfect and may introduce errors. For example, Stable-Zero123~\cite{Zero123} may synthesize incorrect novel-view scenes; DepthAnything~\cite{depthanything} may produce coarse predictions; and multimodal LLMs like LLaVA~\cite{li2024llavanext-strong} may generate wrong answers. However, as shown in our experiments, Eval3D still aligns better with humans than all available automatic 3D evaluations, while being significantly more interpretable.  Furthermore, our proposed framework is designed to support seamless integration of new and improved foundation models. As foundation models continue to improve, we anticipate even stronger alignment with human evaluations.



\vspace{-3mm}
\noindent\paragraph{Broader impact:}
Eval3D can be used to identify and filter out inaccurate and unrealistic 3D assets, and potentially improve existing 3D generation algorithms. It could have a significant impact on the entertainment and robotics industries where high-quality 3D assets are crucial. However, it could also propagate the biases and limitations of foundation models into evaluation metrics. Additionally, it may affect employment within creative fields, as there could be a reduced need for human workers.


\vspace{-3mm}
\noindent\paragraph{Conclusion:} 
We present Eval3D, an interpretable and fine-grained evaluation framework for text/image-to-3D generation algorithms. Eval3D leverages the consistency among various vision and language foundation models to accurately evaluate generated 3D assets along multiple dimensions. It aligns with human judgment better than available alternatives and enables pixel-wise and spatial localization of artifacts in the generated 3D assets. We believe Eval3D can serve as a reliable framework for evaluating, filtering, and potentially improving 3D generation algorithms.

\vspace{-2mm}
\noindent\paragraph{Acknowledgment:} 
The paper was partially supported by a gift from Ai2, an Amazon Science sponsored research grant, NSF CIF 1955864, NSF PHY-2019786, DAF AI Accelerator No. FA8750-19-2-1000, and DARPA TIAMAT program No. HR00112490422. Its contents are solely the responsibility of the authors and do not necessarily represent the official views of DARPA. 

{
    \small
    \bibliographystyle{ieeenat_fullname}
    \bibliography{main}
}

\clearpage

\setlength{\abovedisplayskip}{0pt}
\setlength{\belowdisplayskip}{0pt}
\setlength{\abovedisplayshortskip}{0pt}
\setlength{\belowdisplayshortskip}{0pt}

\appendix

\section{Eval3D Qualitative Analysis}

\begin{figure*}
\footnotesize
    \centering
    \begin{minipage}{0.45\textwidth}
        \centering
        \includegraphics[width=\linewidth,trim={0cm 0.3cm 0cm 0.2cm},clip]{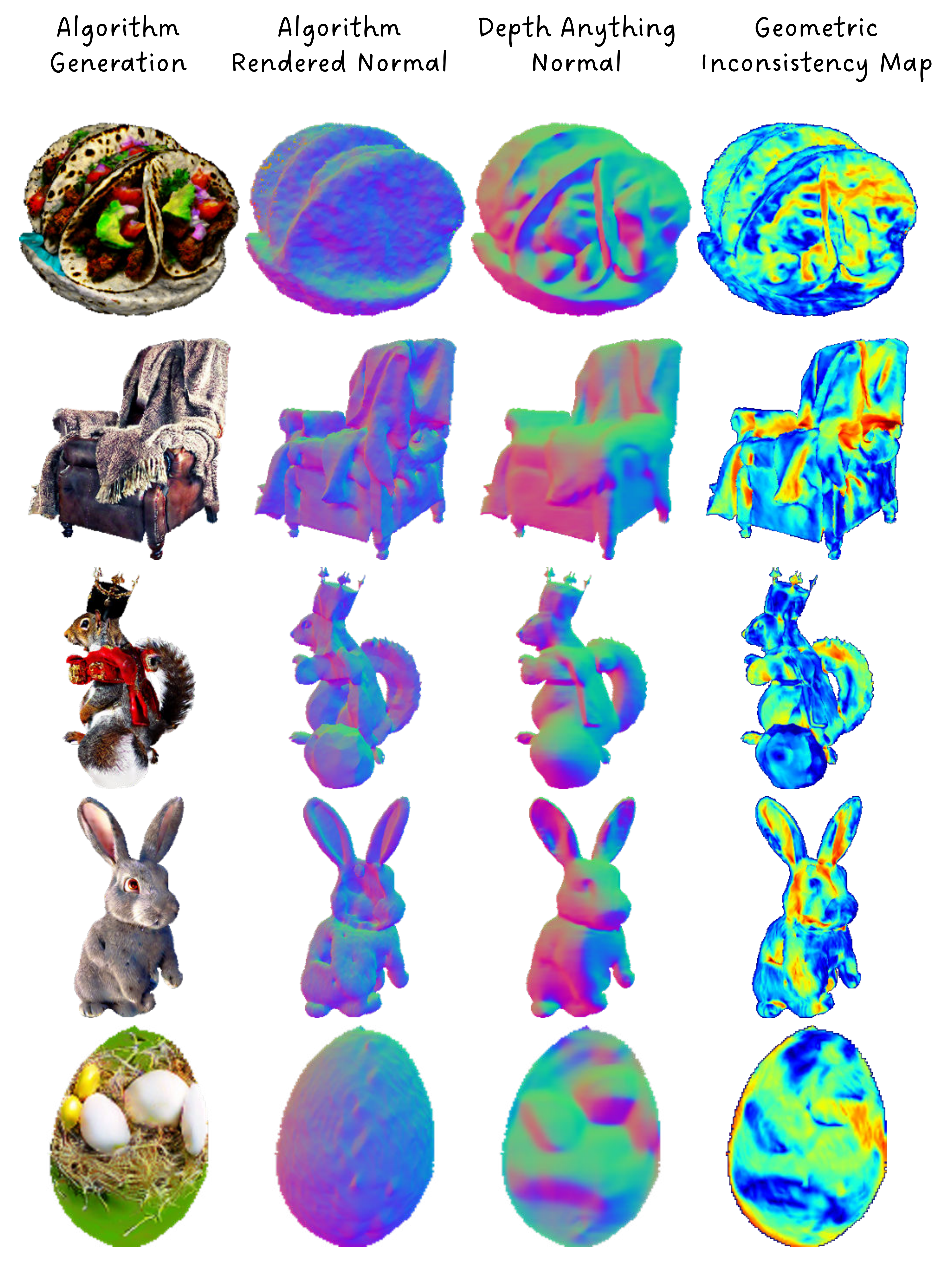}  
    \end{minipage} \hfill
    \begin{minipage}{0.54\textwidth}
        \centering
        \includegraphics[width=\linewidth,trim={0cm 0.3cm 0cm 0.2cm},clip]{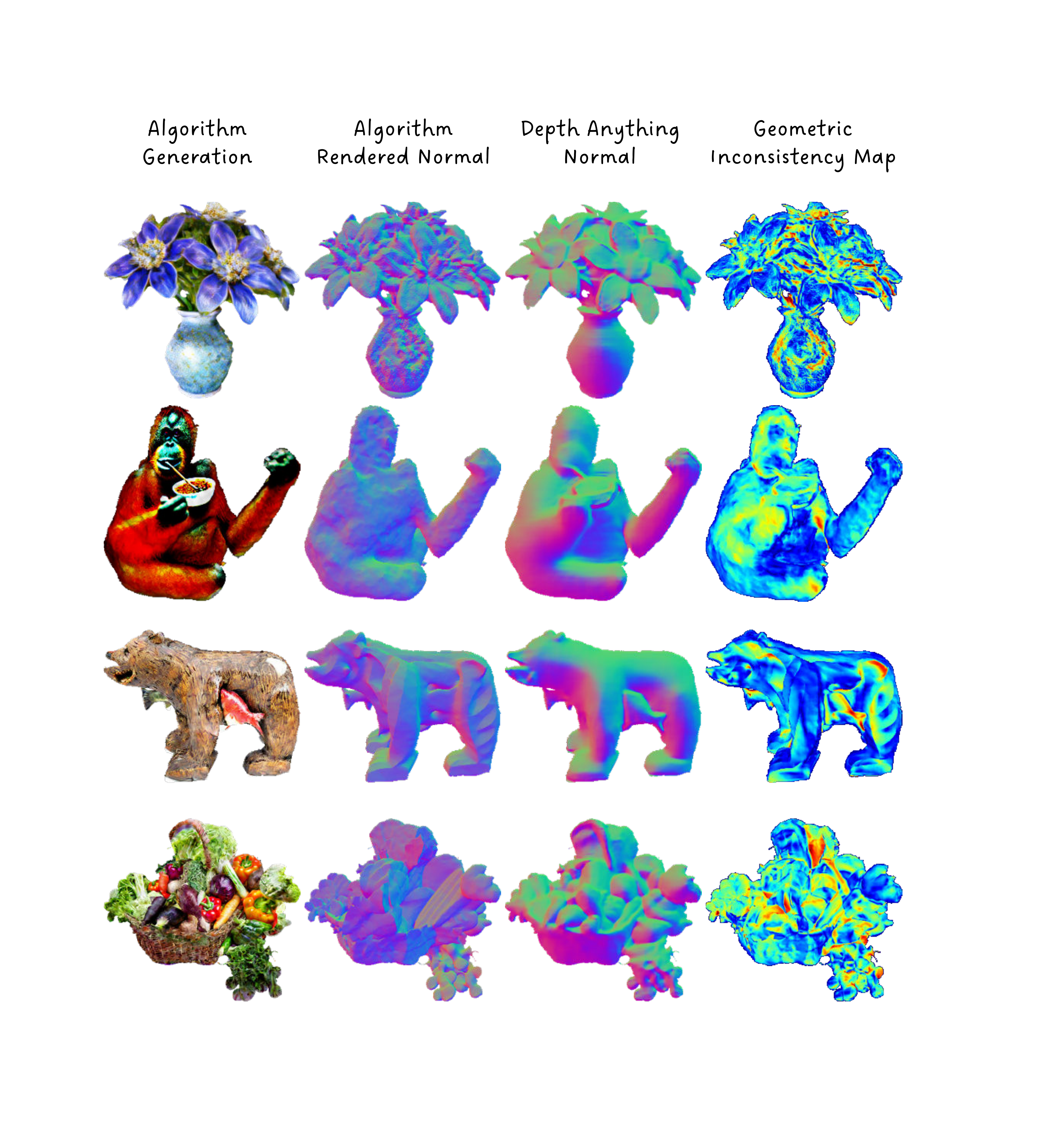}
    \end{minipage} \hfill
    \vspace{-0.1in}
    \caption{\textbf{Geometric Consistency Metric} evaluates texture-geometry alignment by comparing geometrically-rendered normals with image-based Depth-Anything normals. We back-project the consistency estimates onto the 3D mesh to localize 3D artifacts (missing ramen bowl, missing salmon, incorrect vegetable geometry). Bright greenish-red (or cyan) spots = high inconsistency regions , darker-blue regions = consistent areas (we use jet color mode).}
    \label{fig:normal_supp_1}
\end{figure*}

\paragraph{Geometric Consistency:} Fig. \ref{fig:normal_supp_1} illustrate the geometric artifacts identified by our proposed geometric consistency metric. By comparing the image-based normals with the rendered geometric normals, the metric effectively highlights texture-geometry mis-alignments. For instance, see the misalignment between the corresponding texture map and the underlying geometry in the taco fillings, the eggs on the nest, the ramen bowl and the salmon examples of Fig.~\ref{fig:normal_supp_1}. \textbf{Such misalignment are evident when comparing the "Algorithm generation" column with the "Algorithm rendered normal" column.} Among the evaluated text-to-3D algorithms, ProlificDreamer exhibits the most significant texture-geometry misalignments. While MVDream's generations are aesthetically appealing, they sometimes display noisy, non-smooth geometries with holes and cavities, such as the flower pot in the first row (right) of Figure \ref{fig:normal_supp_1}.

\paragraph{Structural Consistency:} Fig. \ref{fig:zero123_supp} qualitatively highlights the structural consistency metric. The first column of the figure shows renderings of the generated 3D assets at $0^\circ, 90^\circ, 180^\circ$ and $270^\circ$. The next two columns present novel view predictions by the zero123 algorithm, using the $0^\circ$ and $90^\circ$ renderings of the generated assets, respectively.

The second and third examples demonstrate faulty generations that suffer from the well-known Janus issue (multi-head reconstructions), where all four viewpoints (in the second and third rows) contain the heads of a chimpanzee and an orange cat, respectively. \textbf{While the text-based 3D generation algorithm struggled, the image-based novel-view synthesis algorithm successfully identified the failure by making multi-view consistent predictions (columns 2, 3).}

Since Zero123 relies on one of the renderings of the generated asset as a reference viewpoint, it can potentially fail to generate consistent predictions if the reference viewpoint contains artifacts or is confusing (e.g., row 3, column 2, generated using the $90^\circ$ rendering of the generated 3D asset in column 1). Rows 1 and 4 showcase examples where good generations are well-aligned with the proposed structural consistency metric, demonstrating qualitative alignment between algorithm generations and zero123 predictions.

\begin{figure}[t]
    \centering \includegraphics[width=\linewidth]{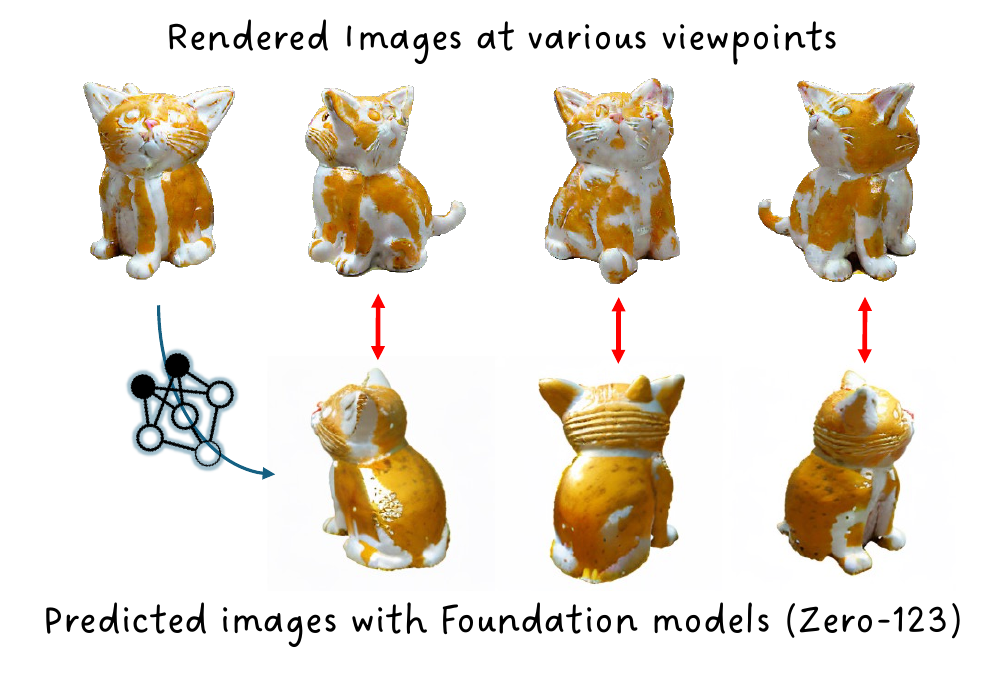}
    \vspace{-4mm}
    \caption{\textbf{Structural consistency:} We compare the rendered images with those predicted by Zero-123 \cite{Zero123}. A structurally coherent object should maintain consistent appearance across different viewpoints, allowing one to predict its appearance from another angle. If the predictions \& renderings differ significantly, it likely indicates an issue. In this figure, generated cat has multiple faces. Since a normal cat only has one face, results from Zero-123 \cite{Zero123} show noticeable inconsistencies, allowing us to localize the structural incoherence.}
    \vspace{-4mm}
    \label{fig:structural-consistency-cat}
\end{figure}

\begin{figure*}[t]
    \centering    
    \vspace{-2mm}
    \includegraphics[width=0.9\linewidth,trim={0cm 0.3cm 0cm 0.2cm},clip]{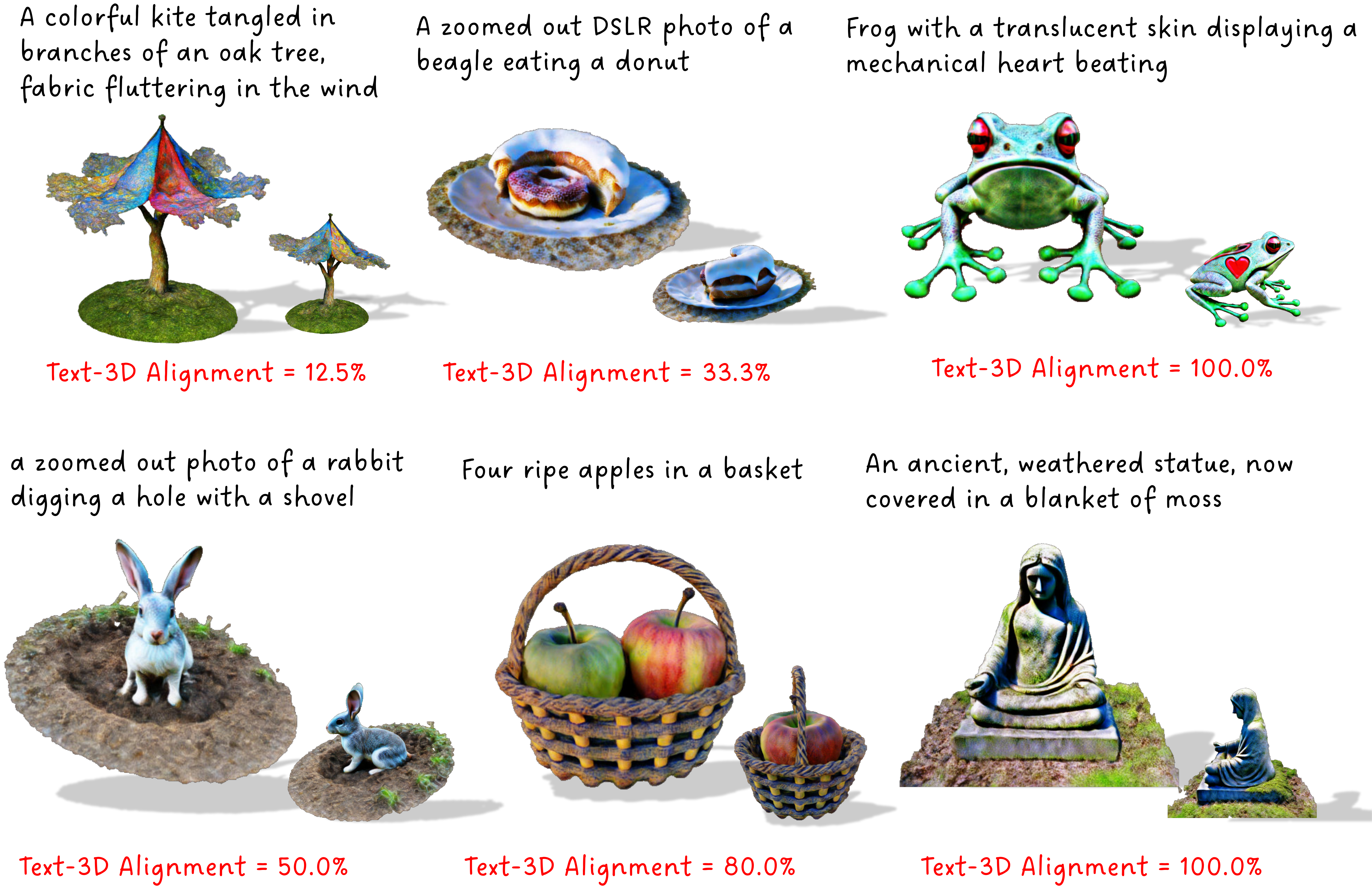}  
    \vspace{-1mm}
    \caption{\textbf{Text-3D Alignment} analyzes how well the generated 3D assets align with user text instructions. We leverage the open-sourced LLaVA model to estimate text-3D alignment. The examples above showcase various scenarios: complete failure of text-3D alignment (first two prompts), missing objects (e.g., the shovel in the fourth example, two apples in the fifth example), and perfect alignment for the last column prompts.}
    \label{fig:tifa_supp}
    \vspace{-5mm}
\end{figure*}

\paragraph{Semantic Consistency:} Fig. \ref{fig:dino_supp} demonstrates our proposed semantic consistency metric. Similar to the geometric consistency metric, the semantic consistency metric is computed directly in the 3D geometric space (as the percentage of semantically consistent mesh vertices), allowing us to localize artifacts in the 3D space. The figure side-by-side compares the generated textured asset with the underlying geometry in the first two columns, while the last column depicts the semantic inconsistency map (bright regions represent mesh vertices with high standard deviation of the back-projected DINO features from neighboring viewpoints). 

The proposed metric effectively localizes artifacts such as Janus issues (multiple dog noses in row 1, multiple pigeon faces in row 4), extraneous geometry (monkey's hand in row 2), noisy texture-geometry alignment (in row 3), and arbitrary floaters in the final row of the figure. Moreover, since DINO features strongly correlate with the semantic class of the underlying image, faulty 3D generations whose semantic interpretation changes with viewpoints are often highlighted by the proposed metric (see Magic3D generation in row 2 of Fig.~\ref{fig:text_3D_comparison_supp_2}, which has a low semantic consistency score of 55.2\%).

\paragraph{Text-3D Alignment:} The alignment of the input text to the generated 3D asset is crucial as it demonstrates how well the 3D generation algorithm follows the input instructions. In Fig.~\ref{fig:tifa_supp}, we highlight the text-3D alignment of several 3D generations. While some of these generations are aesthetically appealing, they fail to satisfy the user's textual instructions (e.g., the first and second generations in the first row). In other cases, certain objects or entities mentioned in the input prompt are missing (e.g., the missing shovel in row 2, column 1, and only two apples generated in column 2). 

Despite demonstrating strong text-3D alignment overall, the generated frog with a mechanical heart (row1, column 3) displays an example where SDS-optimized generations can find unexpected or undesirable ways to satisfy the text prompt.

\section{Text-3D Comparison using Eval3D}
\begin{figure*}[t]
    \centering    
    \vspace{-2mm}
    \includegraphics[width=\textwidth,trim={0cm 0.3cm 0cm 0.2cm},clip]{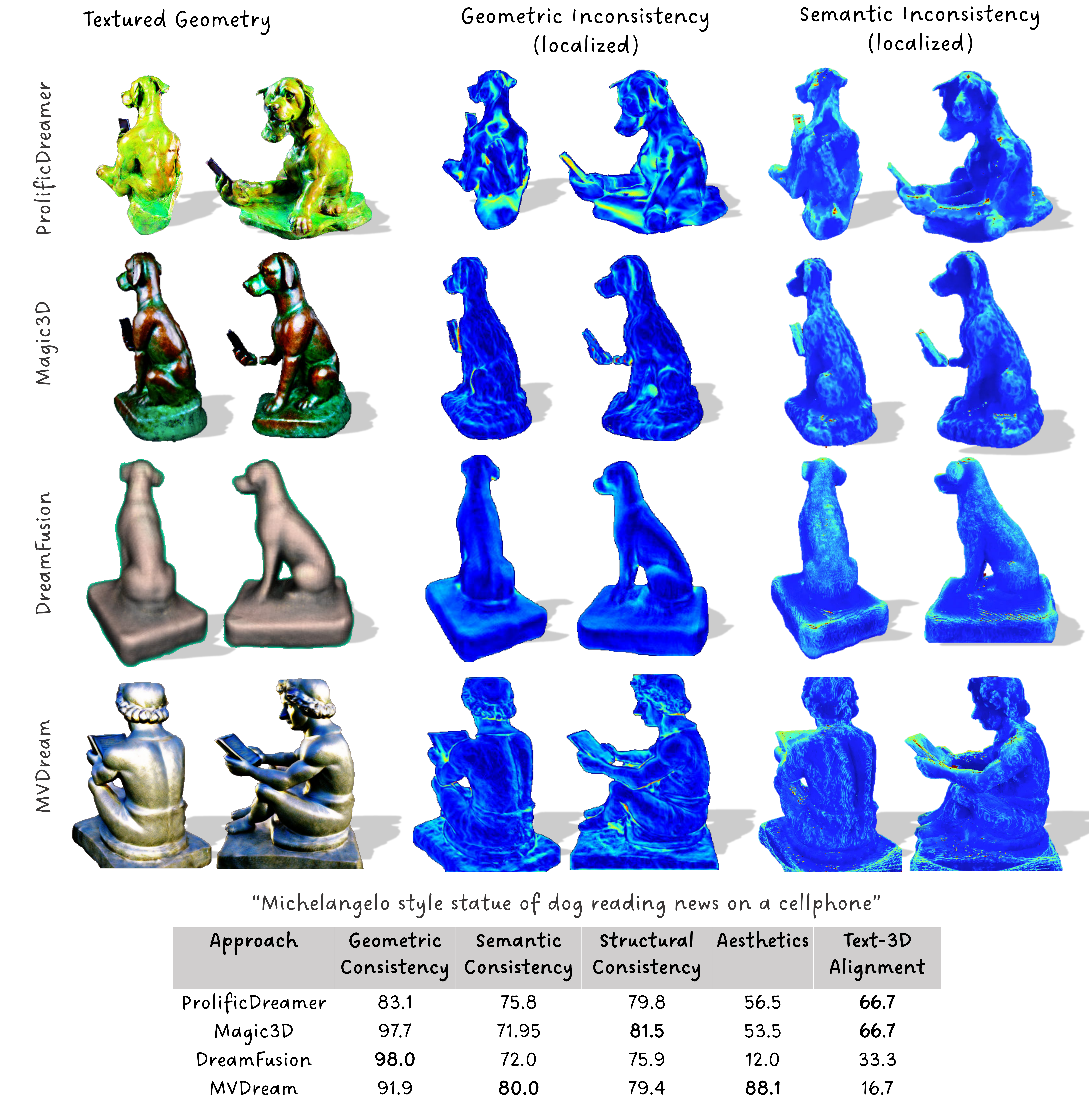}  
    \vspace{-1mm}
    \caption{\textbf{Text-to-3D Generation Comparison}: Magic3D and DreamFusion generate geometrically consistent but overly smooth and simpler geometries that lack aesthetic appeal. MVDream fails miserably to align with this particular prompt, while ProlificDreamer has noticeable localized artifacts in both geometric and semantic inconsistency maps.}

    \label{fig:text_3D_comparison_supp_1}
    \vspace{-5mm}
\end{figure*}

\begin{figure*}[t]
    \centering    
    \vspace{-2mm}
    \includegraphics[width=\textwidth,trim={0cm 0.3cm 0cm 0.2cm},clip]{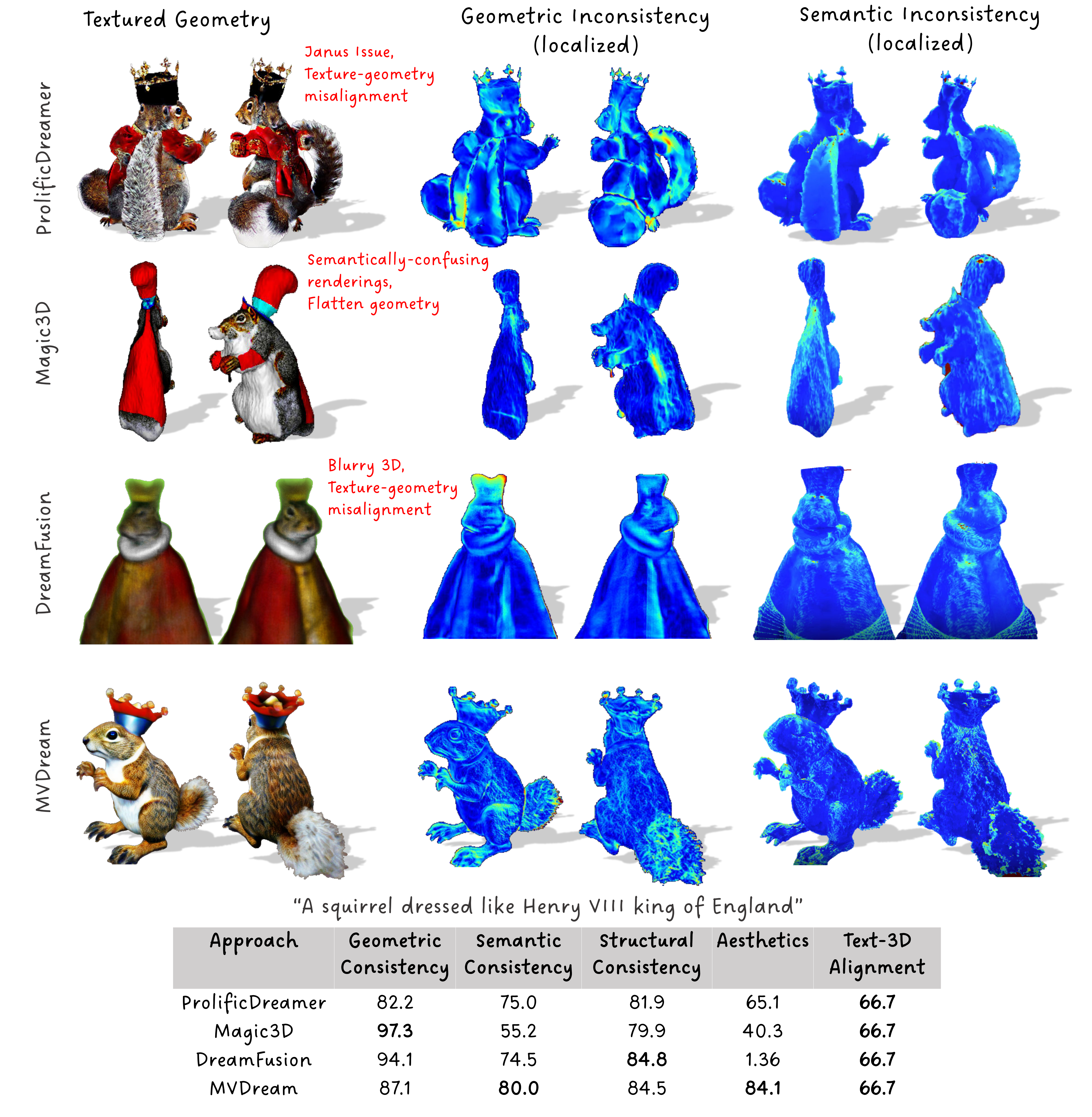}  
    \vspace{-1mm}
    \caption{\textbf{Text-to-3D Generation Comparison}: Magic3D and DreamFusion generate geometrically consistent but overly smooth and simpler geometries that lack aesthetic appeal. Magic3D fails miserably on the semantic consistency metric for this prompt due to semantically confusing renderings from certain viewpoint. Both MVDream and ProlificDreamer exhibit noticeable artifacts in the geometric consistency map for this prompt -- ProlificDreamer shows issues with texture-geometry alignment, while MVDream suffers from noisy generation.}
    \label{fig:text_3D_comparison_supp_2}
\end{figure*}

We compare text-3D generations from multiple algorithms for two prompts in Fig.~\ref{fig:text_3D_comparison_supp_1} and Fig.~\ref{fig:text_3D_comparison_supp_2}. For both the prompts, Magic3D \cite{magic3d} and DreamFusion \cite{dreamfusion} generate overly smooth and simplified geometries, which compromise the aesthetic appeal of the assets. This gets reflected quantitatively in their high geometric consistency scores and qualitatively in the corresponding geometric inconsistency maps as shown in the second column (bright spots indicate higher inconsistency, while darker-blue regions indicate lower error).

For each algorithm's generation in both figures, the semantic consistency metric effectively localizes artifacts in the 3D space. For example, ProlificDreamer exhibits Janus issues, Magic3D shows semantically confusing renderings of relatively thin geometries, and DreamFusion displays low texture-geometry alignment in its blurry reconstructions. Finally, MVDream completely fails to satisfy the text prompt in the first figure, generating a Michelangelo-style human instead of a dog.

\section{Eval3D Failure Cases}

\begin{figure*}
\footnotesize
    \centering
    \begin{minipage}{0.5\textwidth}
        \centering
        \includegraphics[width=\textwidth,trim={0cm 0.3cm 0cm 0.2cm},clip]{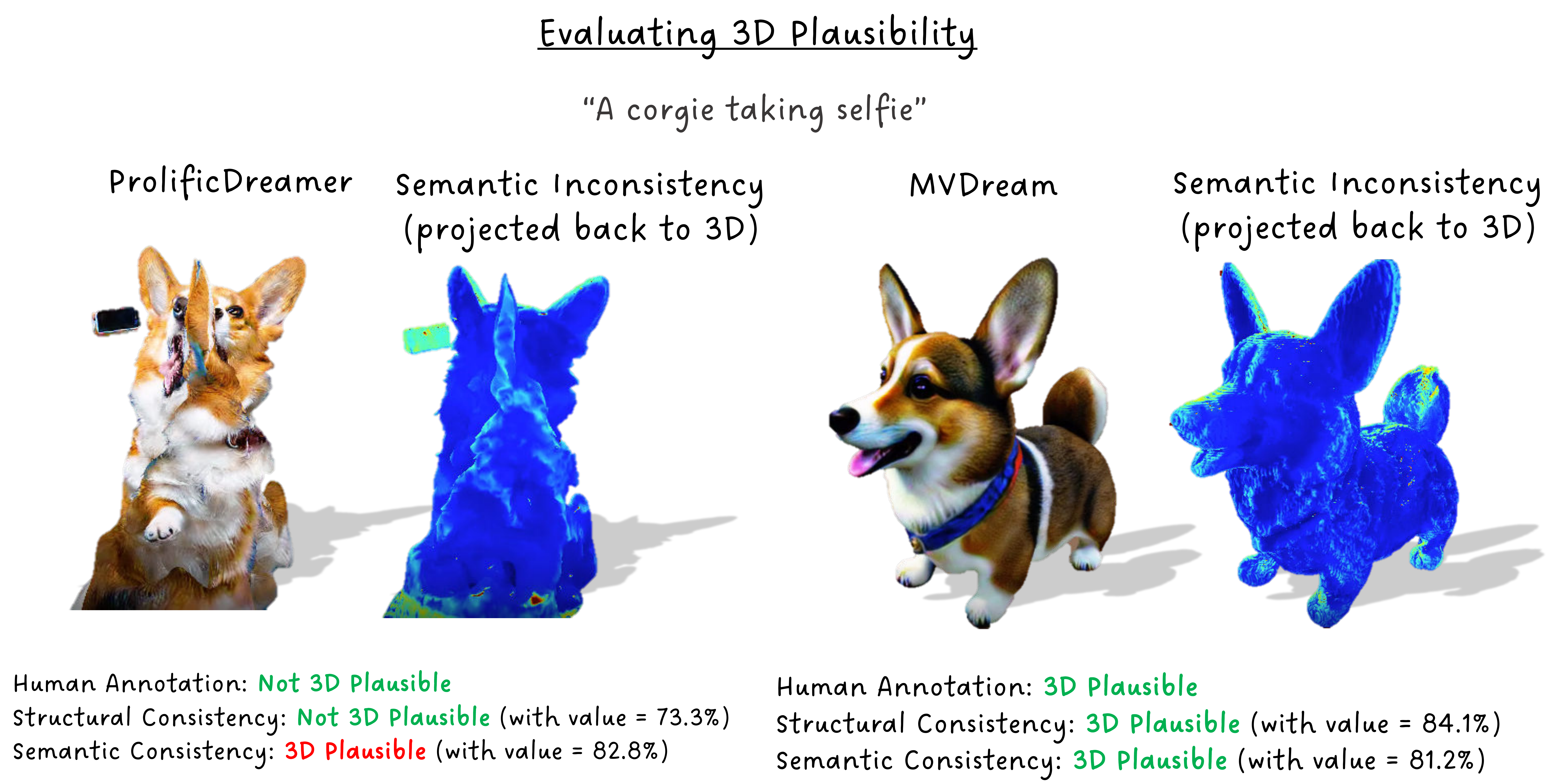}   
    \end{minipage} \hfill
    \begin{minipage}{0.49\textwidth}
        \centering
        \includegraphics[width=\textwidth,trim={0cm 0.3cm 0cm 0.2cm},clip]{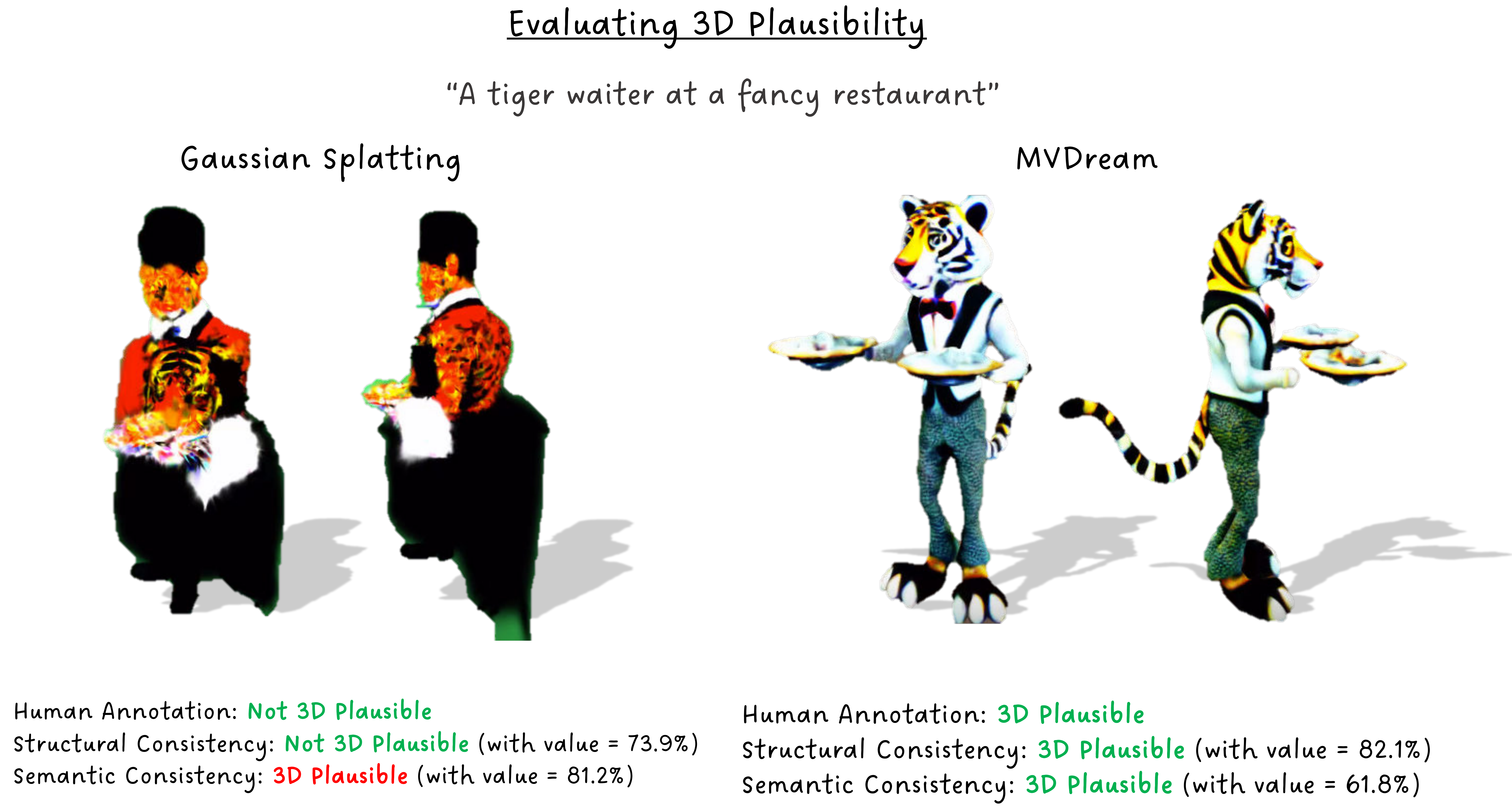}
    \end{minipage} \hfill
    \caption{\textbf{Eval3D Semantic Consistency Failure Cases} Potential Reasons for failures -- Artifacts (intersection of multiple faces) being localized to very thin regions; occlusion of geometric structures for certain viewpoints, making the overall geometry's semantic interpretation ambiguous; noisy gaussian splatting structures being too OOD for DINO model. (Zoom-in for best visualization)}
    \label{fig:failure_supp_1}
\end{figure*}

While the proposed 3D evaluation metrics align better than current alternatives and offer an interpretable, fine-grained solution to 3D evaluation, there are still some failure cases that warrant future investigation. Fig.~\ref{fig:failure_supp_1} demonstrates two such failure cases for semantic consistency metric.

In Fig.~\ref{fig:failure_supp_1}, the semantic consistency metric successfully localizes artifacts for both ProlificDreamer (regions with Janus issues, i.e., intersecting multiple heads) and MVDream (regions with poor texture-geometry alignment due to dog fur-like texture). However, the quantitative semantic consistency score for ProlificDreamer fails to align with human judgment for this particular prompt (while overall Prolificdreamer-Human alignment using semantic consistency metric equals 63\%). We hypothesize that this is because the artifact is localized to a relatively thin region of semantic confusion where multiple heads intersect, thereby affecting the metric value less significantly. Another cause of failure is the occlusion of geometric parts at certain viewpoints, which potentially leads to different semantic interpretations from those viewpoints, resulting in higher semantic inconsistency values for such vertices (MVDream's dog nose).

Fig.~\ref{fig:failure_supp_1} (right) highlights a particular scenario where a Gaussian-splatting-based algorithm generates relatively noisy and potentially out-of-distribution textures, making it difficult for the DinoV2 model to derive any meaningful semantic interpretation from the corresponding renderings.

\section{Experimental Details}

\subsection{Training Details}
We evaluated six text-to-3D and two image-to-3D algorithms for Eval3D benchmark. For all algorithms, we leveraged code and implementations from the Threestudio project \cite{threestudio2023} (or its extensions). Note that Threestudio's implementations have some differences from the original works. We refer readers to the "Notable differences from the paper" sections corresponding to each algorithm on the Threestudio project page. All our experiments were done on NVIDIA A40 GPUs. Most compute cost are spent on generating 3D assets for benchmarking, and the compute cost varies for different generation methods, ranging from 0.3 GPU hours (for DreamFusion~\cite{dreamfusion}) to 6 GPU hours (for ProlificDreamer~\cite{prolificdreamer}). For evaluation, after rendering the mesh and 120 RGB images, the majority of compute cost is spend on running the foundation models, including LLaVA-7B~\cite{li2024llavanext-strong}, DINOv2~\cite{DINOv2}, and Stable-Zero123~\cite{Stable-Zero123}. Altogether, when run parallelly, these processes take about 5 minutes per 3D asset.

\subsection{Hyper-parameter Selection}
We follow standard practices by normalizing the object within $[-1, 1]$, sampling cameras around it, and rendering $256\times256$ images. 
Foundation models like DINO are robust to varying resolutions. The threshold for each metric is determined using a hold-out validation set, and the number of views is selected empirically.

\subsection{Mesh Extraction and Vertex Visibility} We use marching cubes or marching tetrahedra to extract meshes from the learned density or tetrahedral fields (in Magic3D). 
While this process may introduce mesh artifacts, we find that, in practice, it has a limited effect on the semantic consistency metric compared to inconsistencies caused by other factors, such as the Janus effect. 
We only consider vertices that are visible from at least five viewpoints. Vertex visibility can be determined by rendering the mesh via rasterization. This helps avoid labeling spurious vertices (which exhibit low DINO variance due to being visible in very few frames) as consistent vertices. 

\subsection{Design of Structural Consistency Metric} 
Based on the observation that 3D generation algorithms typically perform well under small viewpoint shifts but struggle with significant changes, we evaluate every $90^\circ$ to balance computational cost and effectiveness. The camera is positioned at a $70^\circ$ elevation with varying azimuth angles to fully capture the top and front viewpoints. Bottom viewpoints are excluded, as they are often planar and semantically simpler, allowing us to conserve computation.
To aggregate the scores across different inputs (\textit{e.g.}, $0^\circ$ vs. $90^\circ$), we experimented with both \texttt{max} and \texttt{average}. We find that while Zero123 is strong, it is not perfect and may perform worse on some input images due to noise. Taking the \texttt{max} enhances robustness to this variability. 

\section{Eval3D Benchmark}

Here we provide more details about the Eval3D benchmark beyond what covered in the main paper experiments section. Figure~\ref{fig:eval3d_stat} contains the statistics calculated from these scene graphs. The left histogram shows the number of prompts that have a certain number of entities. The right shows the statistics of the total number of semantic elements in a scene graph. Figure~\ref{fig:eval3d_scenegraph} illustrates the frequencies of each semantic element type. We also give the sub-category for entity and attribute. Here ``Entity whole'' is the major objects (e.g., ``horse''), and ``Entity part'' is part of a major object (e.g., ``saddle on a horse'').

\subsection{Human Annotations}
Eval3D is annotated by 10 computer vision graduate students who have been well-trained to do the task. We strictly follow their institutions' rules during training and annotation. All annotators help for free and we really appreciate their effort. The major potential participation risk is that some 3D assets are so low-quality that they might be visually disturbing to humans. We collect annotations for 160 prompts on 6 models and 4 evaluation criteria. The annotation guideline is as follows:

\paragraph{Geometric Consistency:} For each text prompt, we show annotators videos of RGB and rendered surface normals, placed side-by-side for all six algorithms. The annotator needs to give each 3D asset a score between 0--9, focusing on the \textit{texture alignment with the surface normal}. We only use the scores for pairwise comparison. Ties are allowed. The annotation interface is shown in Fig.~\ref{fig:geo_consis_interface}. The inter-annotator agreement, in terms of pairwise agreement (i.e. the probability that both annotators believe a 3D asset is better than the other), is 86.8\%.\\
\begin{figure*}[t]
    \centering \includegraphics[width=0.75\linewidth]{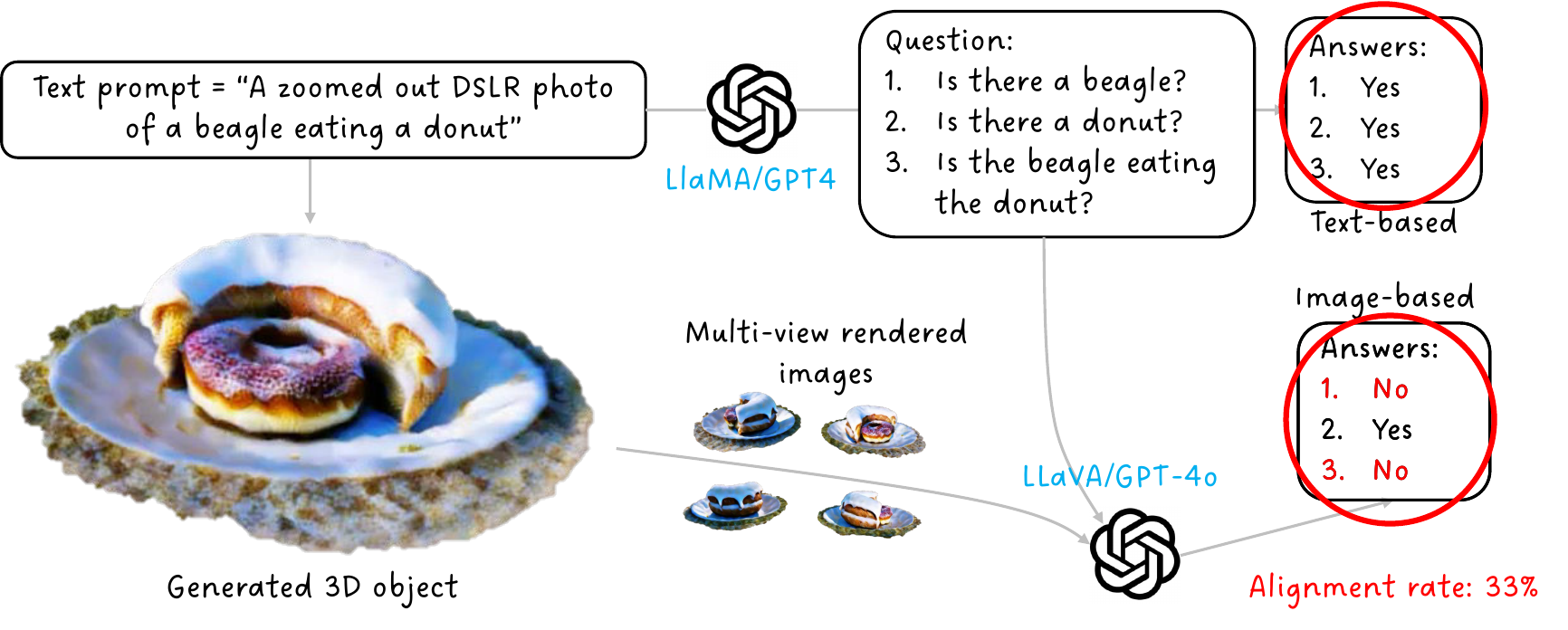}
    \caption{\textbf{Illustration of Eval3D Text-3D alignment pipeline.}}
    \label{fig:tifa3D-pipeline}
\end{figure*}

\begin{table*}[t]
\footnotesize
    \centering
    \begin{minipage}{0.62\textwidth}
        \centering
        \includegraphics[width=\linewidth]{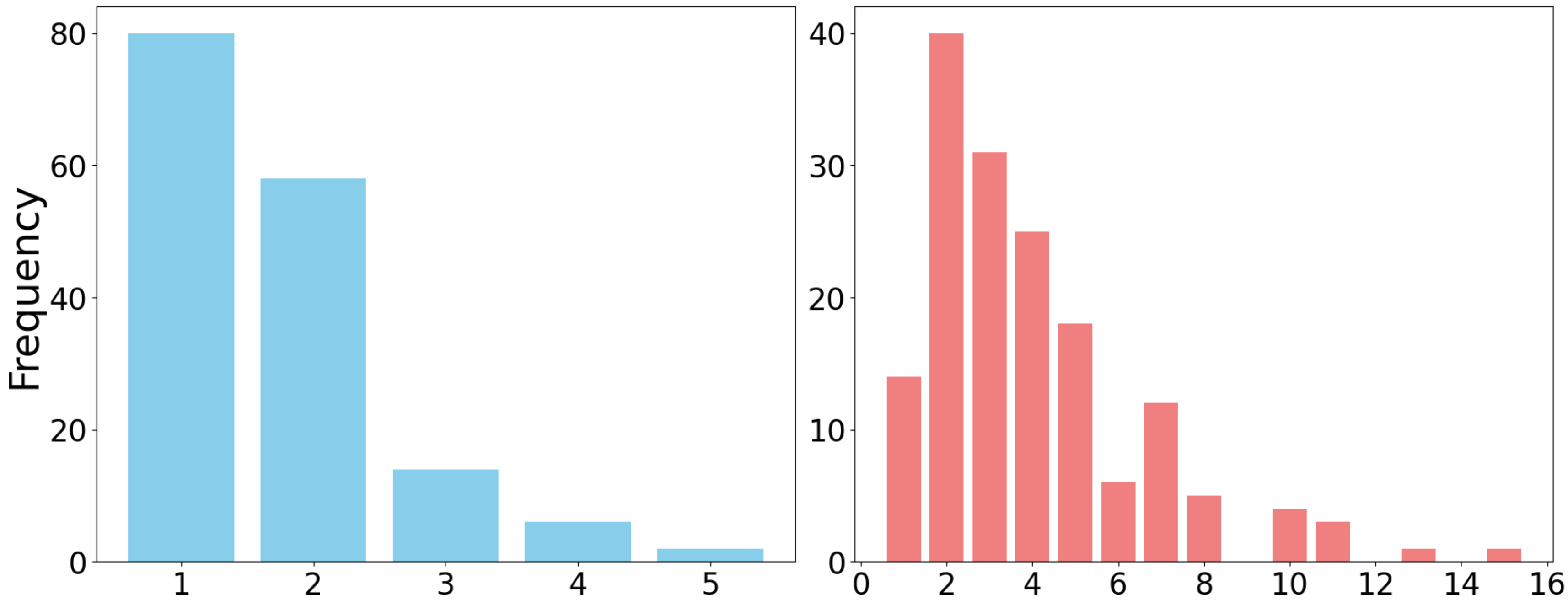}
        \captionof{figure}{\textbf{Statistics of prompts in Eval3D Benchmark }\textbf{Left:} The number of entities in a prompt. \textbf{Right:} Number of semantic elements in a Prompt.}
        \label{fig:eval3d_stat}
    \end{minipage} \hfill
    \begin{minipage}{0.34\textwidth}
        \centering
        \includegraphics[width=\linewidth]{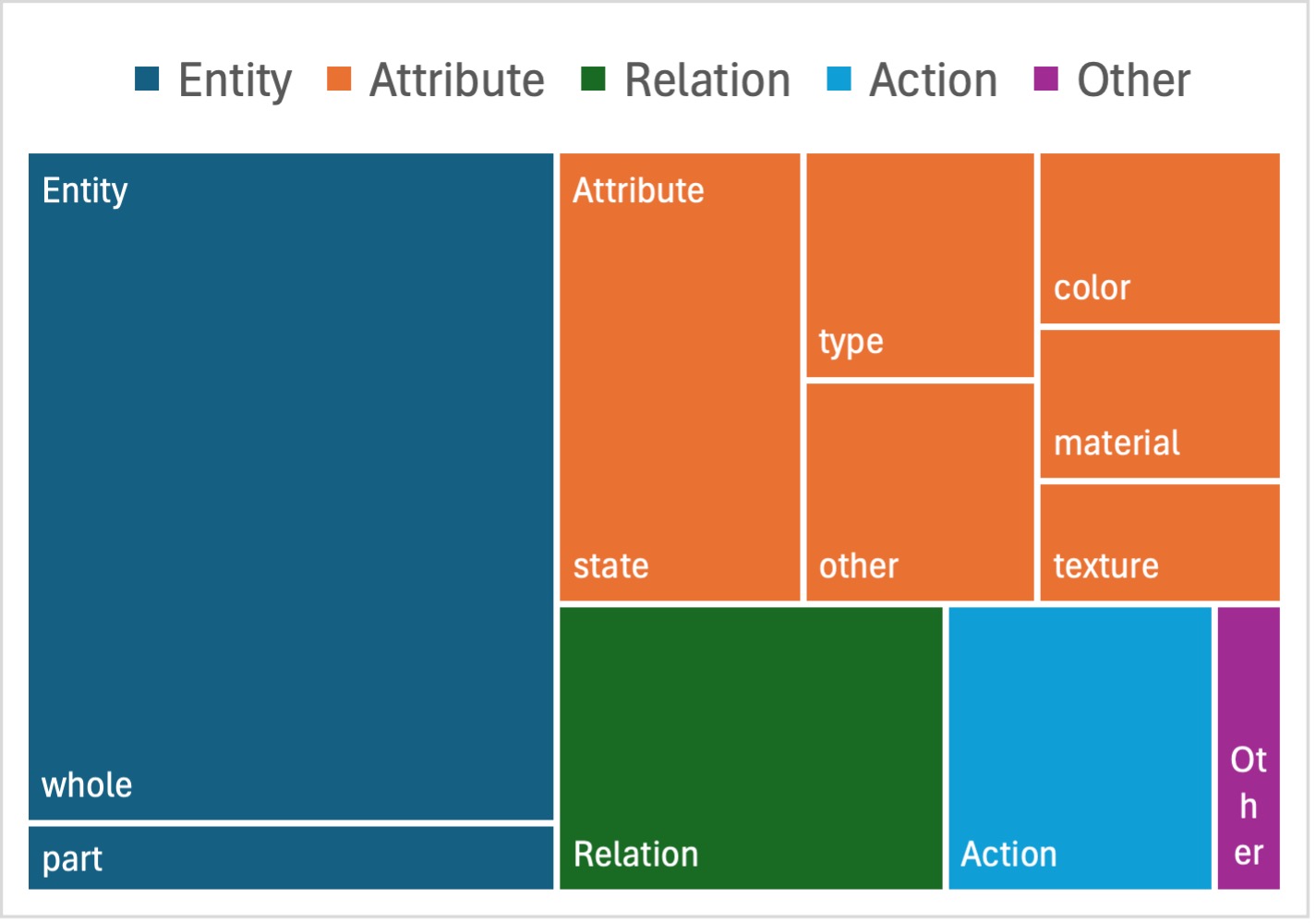}
        \captionof{figure}{\textbf{Frequencies of each type of semantic element in the Eval3D scene graphs.}}
        \label{fig:eval3d_scenegraph}
    \end{minipage} \hfill
    \vspace{2mm}
\end{table*}

\begin{figure*}[t]
    \centering    
    \includegraphics[width=0.8\textwidth,trim={0cm 0.3cm 0cm 0.2cm},clip]{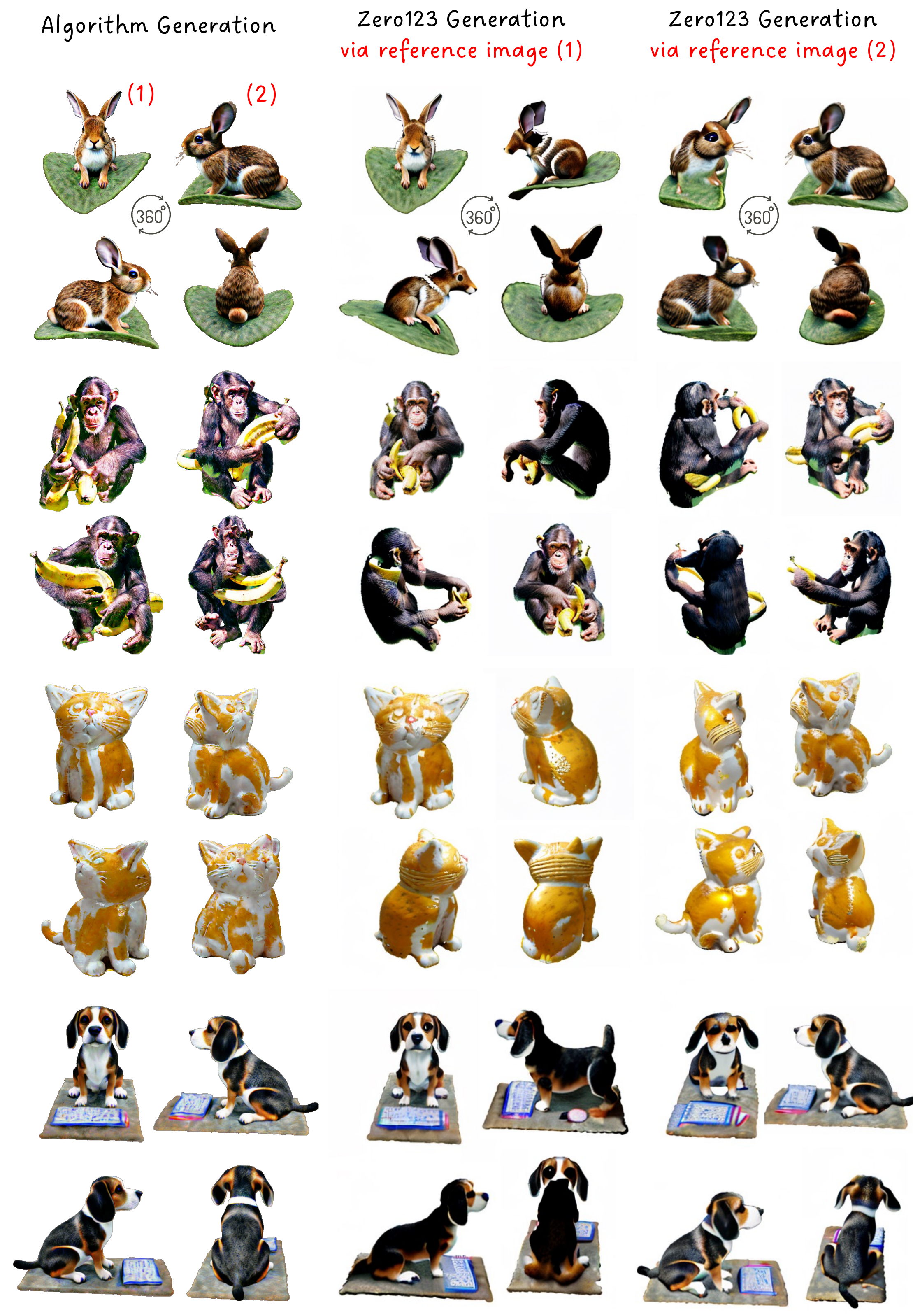}  
    \vspace{-1mm}
    \caption{\textbf{Structural Consistency} measures overall 3D plausibility by comparing (via Dreamsim) the text-based 3D asset renderings with predictions from the image-based novel view synthesis algorithm, Zero123 (i.e., comparing column 1 with columns 2 and 3). The middle rows highlight faulty generations with Janus issues, while the remaining rows showcase multi-view consistent generations.}

    \label{fig:zero123_supp}
\end{figure*}

\begin{figure*}[t]
    \centering    
    \vspace{-2mm}
    \includegraphics[width=0.8\linewidth,trim={0cm 0.3cm 0cm 0.2cm},clip]{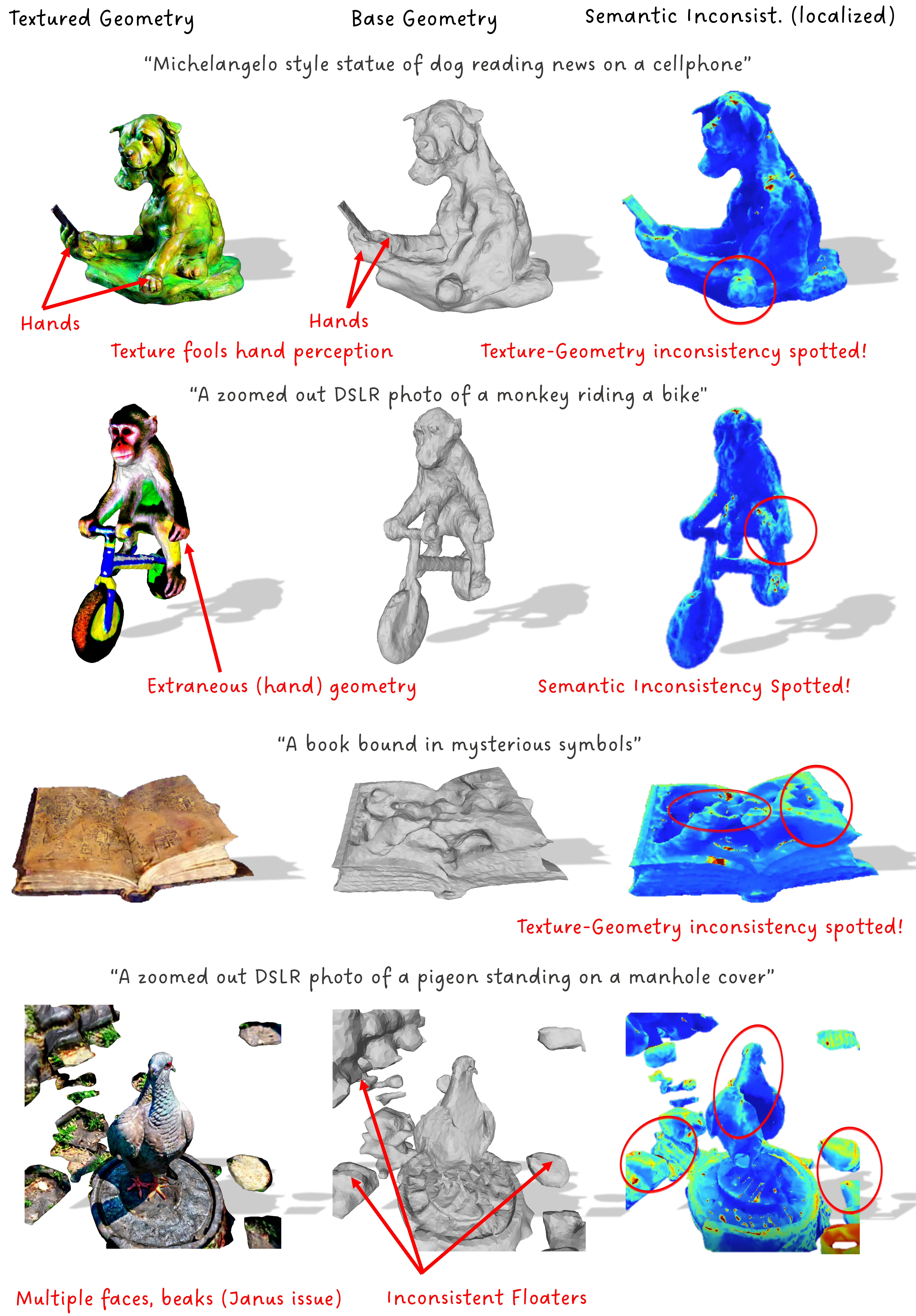}  
    \vspace{-1mm}
    \caption{\textbf{Semantic Consistency} leverages the DinoV2 foundational model to measure the multi-view semantic consistency of each mesh vertex. We showcase various scenarios that could lead to multi-view semantic confusion, such as Janus issues, extraneous geometry, incorrect texture-geometry alignment, and generated floaters. Incorrect texture-geometry alignment will lead to incorrect back-projection of 2D Dino features onto 3D points, leading to high inconsistency.}
    \label{fig:dino_supp}
    \vspace{-5mm}
\end{figure*}

\paragraph{Semantic and Structural Consistency:} For each 3D asset, we ask the annotator to judge its structural consistency or 3D plausibility by showing then a rendered RGB video. We educate them about artifacts like Janus issues (corgi in Fig.~\ref{fig:zero123_supp}), fluctuating semantics (Fig.~\ref{fig:dino_supp}), arbitrary/incorrectly rendered novel views (Fig.~\ref{fig:zero123_supp}).
The annotator chooses among four options: \textit{yes}, \textit{uncertain yes}, \textit{uncertain no}, and \textit{no}. The annotation interface is shown in Fig.~\ref{fig:geo_consis_interface}. We compute the inter-annotator agreement in two ways: (1) excluding the examples where annotators answer ``uncertain yes'' or ``uncertain no'', then the pairwise agreement between annotators is 97.2\%. (2) if we consider ``yes'' and ``uncertain yes'' as one class, ``no'' and ``uncertain no'' as the other class, then the agreement is 83.1\%.

\begin{figure}[t]
    \centering
    \includegraphics[width=0.9\linewidth]{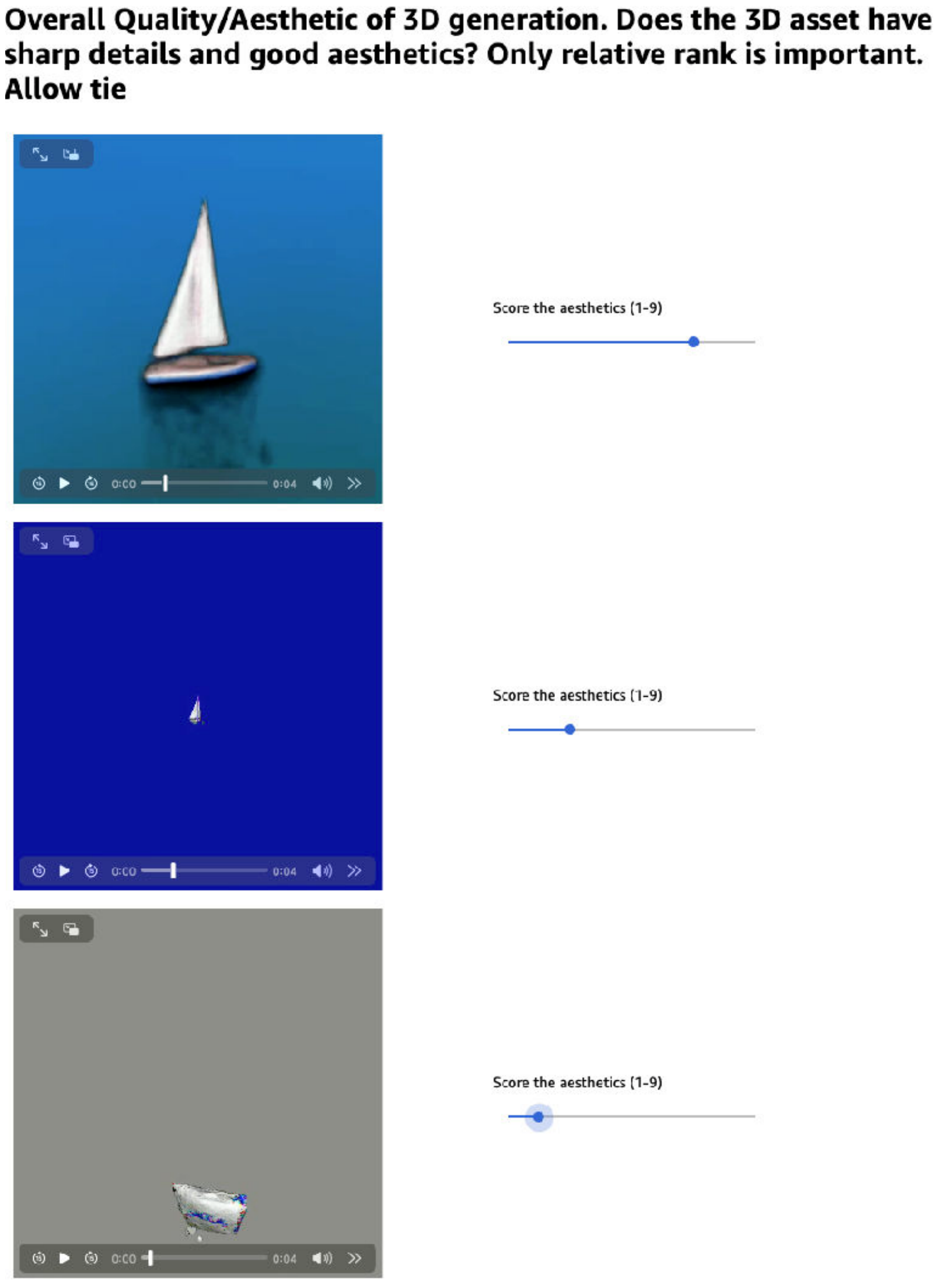}
    \caption{\textbf{Aesthetics Annotation Interface:} For each prompt, we show the RGB videos of the assets generated by all six 3D generation models. We only show 3 of them here.}
    \label{fig:aesthetics_interface}
\end{figure}

\paragraph{Aesthetics:} We use the same annotation template as for geometric consistency. For each text prompt, we show rendered RGB videos of 3D assets of all six algorithms. The annotator needs to rank the generations based on \textit{whether a 3D asset is aesthetically pleasing, containing sharp, natural, vivid, bright, and high-resolution textures.} Ties are allowed. The interface is illustrated in Fig.~\ref{fig:aesthetics_interface}. The inter-annotator agreement is 83.8\%.

\begin{figure}[t]
    \centering
    \includegraphics[width=0.9\linewidth]{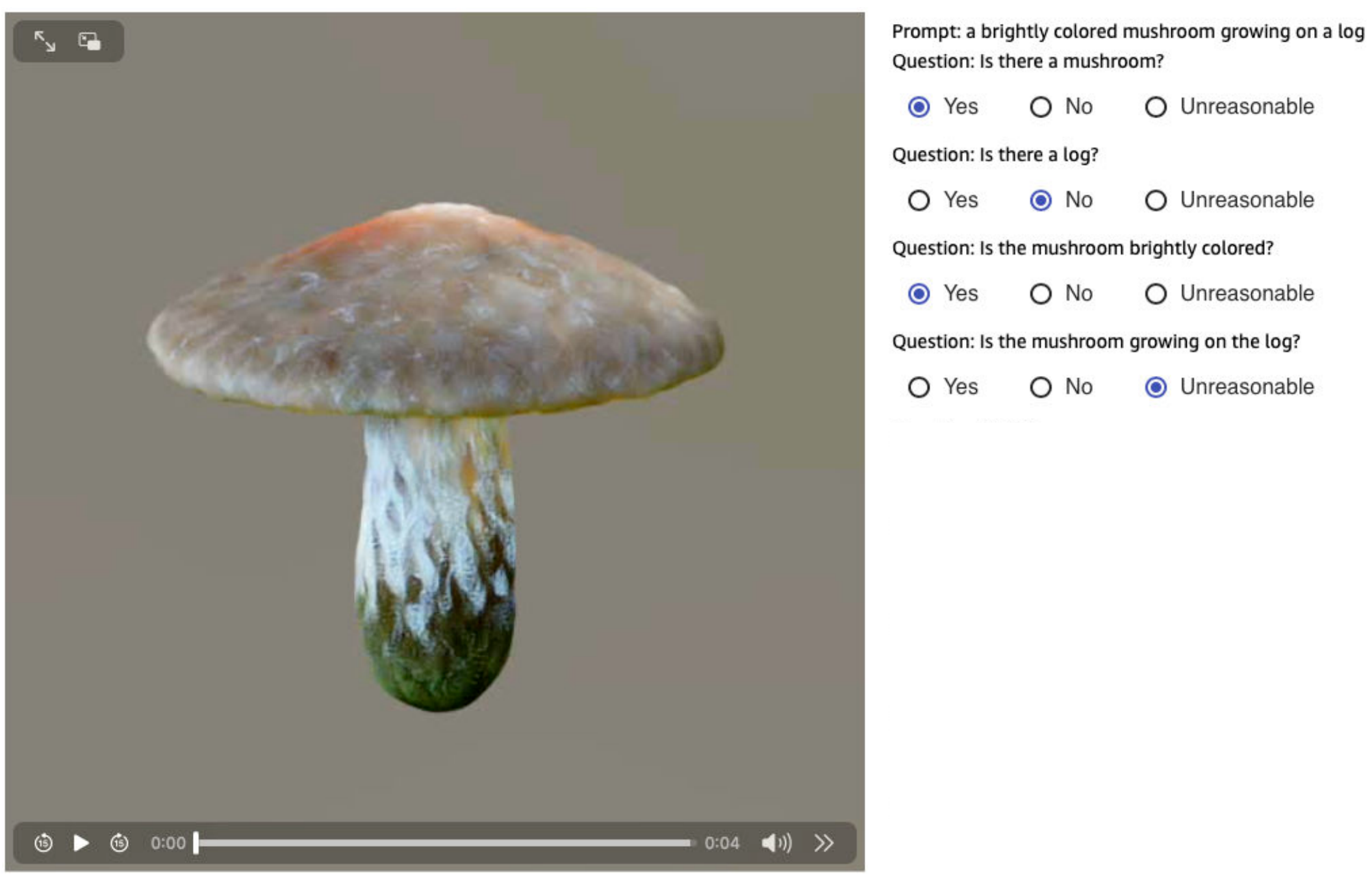}
    \caption{\textbf{Text-3D Alignment Annotaiton Interface.}}
    \vspace{-2mm}
    \label{fig:tifa_interface}
\end{figure}

\paragraph{Text-3D Alignment:} We adopt the annotation template of previous text-to-image evaluation works~\cite{JaeminCho2024, wiles2024revisiting} and evaluate RGB videos instead of images. For each 3D asset, We ask annotators to answer a series of questions generated automatically from the text prompt, as discussed in~\cite{JaeminCho2024}. The annotators perform multiple-choice video question answering. They can also mark a question as \textit{unreasonable} in case they believe the question is not reasonable. 
The annotation interface is shown in Fig.~\ref{fig:tifa_interface}. We compute a score for each 3D asset by counting how many questions have been answered ``yes''. The inter-annotator agreement, in terms of pairwise comparison between 3D assets, is 95.4\%.

\textbf{Note on computing automatic evaluations' alignment with human.} For geometric consistency, aesthetics, and text-3D alignment, we compute evaluation metrics' alignment with humans using the same way as we compute the inter-annotator agreement, i.e. pairwise comparison agreement. For semantic and structural consistency, humans annotate ``yes'' ``no'' while the automatic evaluation gives a continuous value. We process the automatic evaluation by finding a threshold to divide its scores into two classes. For all evaluation metrics, we report the maximum value of human alignment given all possible thresholds. For Eval3D, the threshold for structural consistency is 75.8\%; for semantic consistency it is 63.3\%.

\clearpage
\newpage 
\begin{figure}[t]
    \centering
    \includegraphics[width=\linewidth]{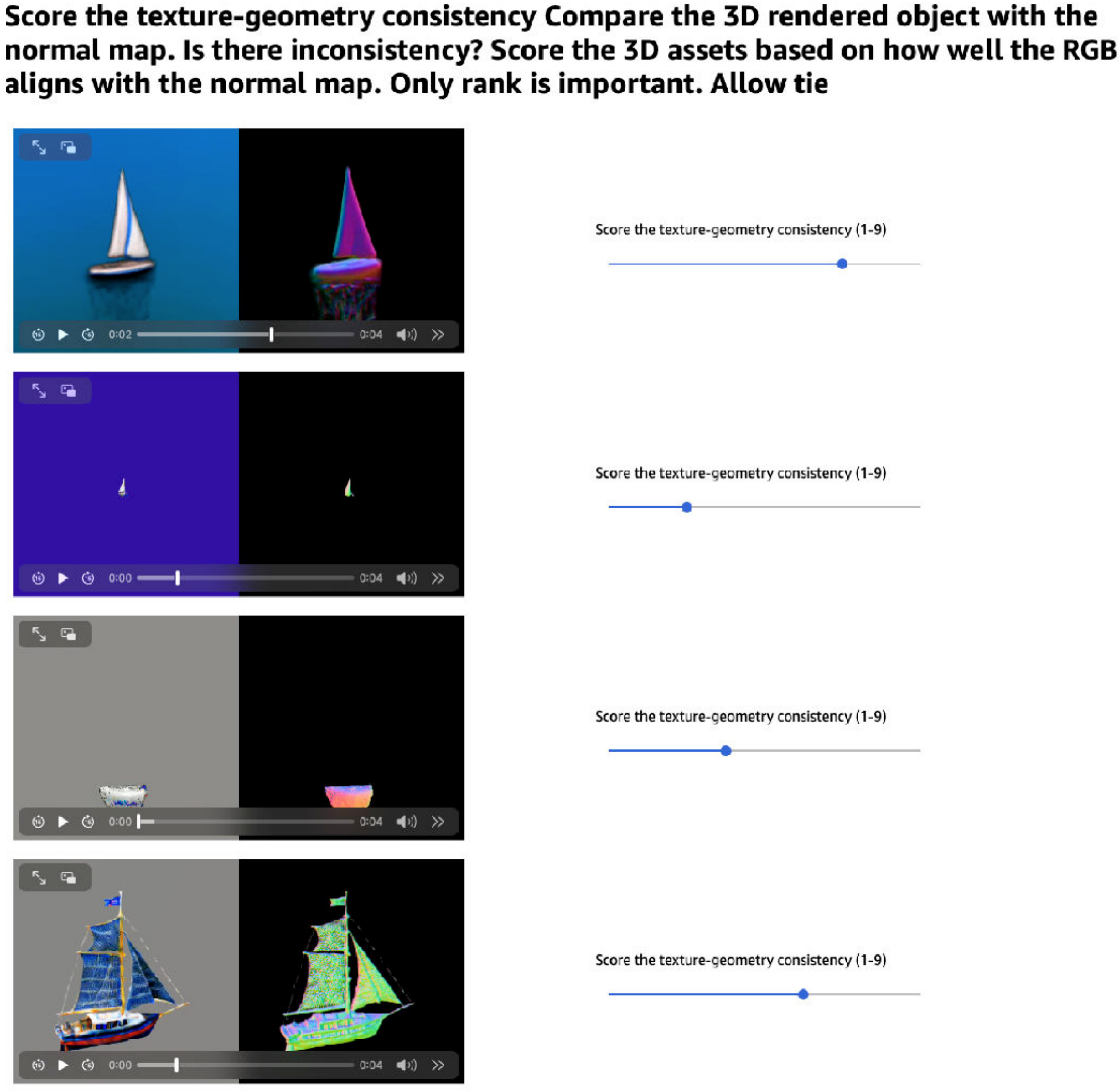}
    \caption{\textbf{Geometric Consistency Annotation Interface: }For each prompt, we show RGB \& normal map videos of the assets generated by all six 3D generation models. We only show 4 of them here.}
    \label{fig:geo_consis_interface}
\end{figure}

\begin{figure}[t]
    \centering
    \includegraphics[width=\linewidth]{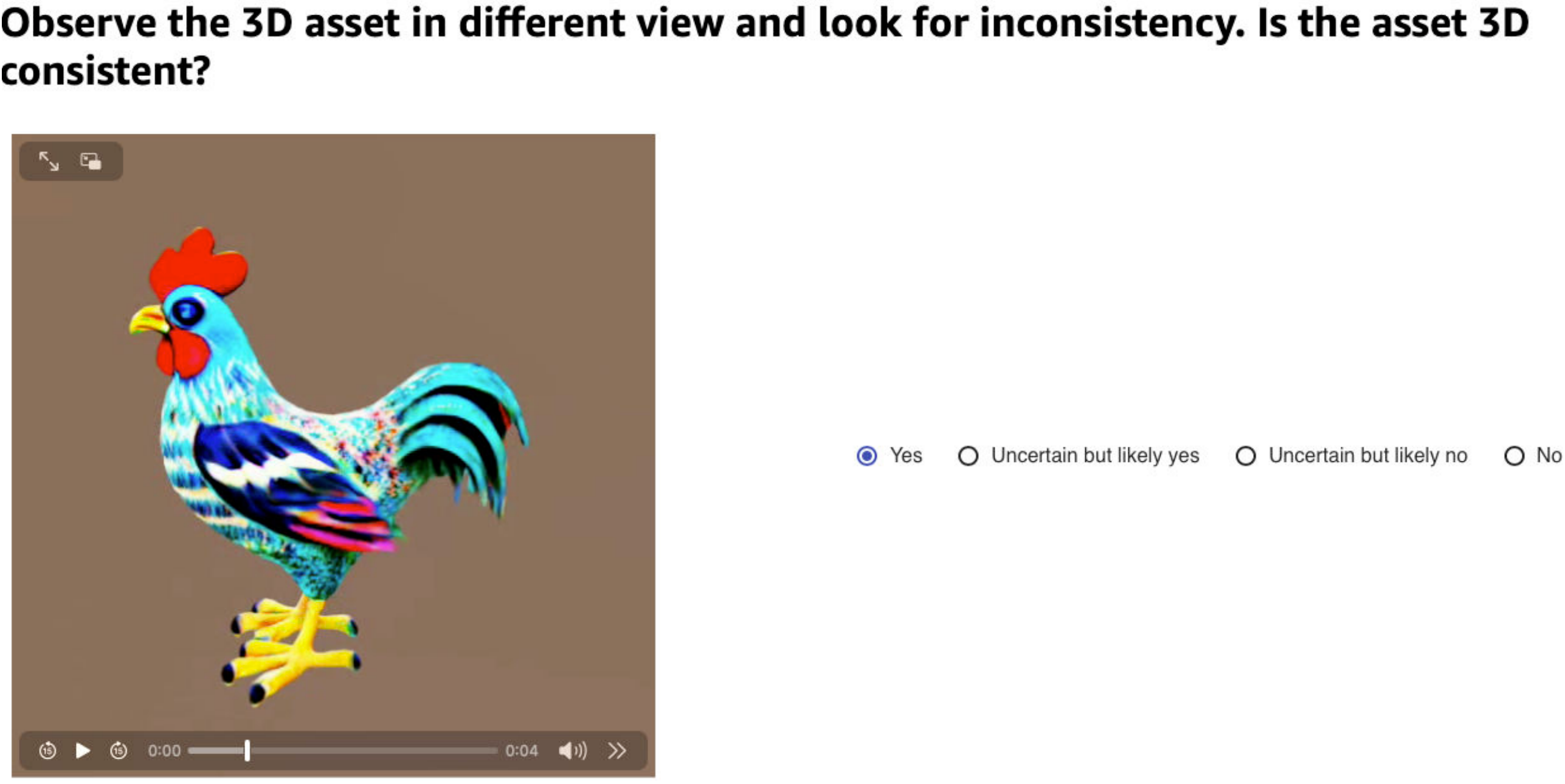}
    \caption{\textbf{Structural \& Semantic Consis. Annotation Interface.}}
    \vspace{-4mm}
    \label{fig:geo_consis_interface}
\end{figure}

\end{document}